\documentclass[10pt,twocolumn,letterpaper]{article}

\usepackage{iccv}
\usepackage{times}
\usepackage{epsfig}
\usepackage{graphicx}
\usepackage{amsmath}
\usepackage{amssymb}
\usepackage{microtype}
\usepackage{paralist}
\usepackage{enumerate}
\usepackage{amsthm}
\usepackage{multirow}
\usepackage{enumitem}
\usepackage{hhline}% http://ctan.org/pkg/hhline
\usepackage{array}
\usepackage{wrapfig}
\usepackage{xfrac}

\newenvironment{tight_itemize}{
\begin{itemize}[leftmargin=10pt]
  \setlength{\topsep}{0pt}
  \setlength{\itemsep}{0pt}
  \setlength{\parskip}{0pt}
  \setlength{\parsep}{0pt}
}{\end{itemize}}

\usepackage[pagebackref=true,breaklinks=true,letterpaper=true,colorlinks,bookmarks=false]{hyperref}

\iccvfinalcopy % *** Uncomment this line for the final submission

\def\iccvPaperID{1499} % *** Enter the ICCV Paper ID here

\ificcvfinal\fi
\begin{document}

%%%%%%%%% TITLE
\title{Unsupervised Domain Adaptation for Face Recognition in Unlabeled Videos}

\author{Kihyuk Sohn$^1$
\hspace{0.08in} Sifei Liu$^{2}$
%University of California, Merced^2\\
\hspace{0.08in} Guangyu Zhong$^3$
\hspace{0.08in} Xiang Yu$^1$
\hspace{0.08in} Ming-Hsuan Yang$^2$
\hspace{0.08in} Manmohan Chandraker$^{1,4}$
\vspace{1mm} \\
\hspace{0.15in} $^1$NEC Labs America
\hspace{0.15in} $^{2}$UC Merced
\hspace{0.15in} $^{3}$Dalian University of Technology
\hspace{0.15in} $^{4}$UC San Diego\\
}
\maketitle
\thispagestyle{empty}

%%%%%%%%% ABSTRACT
\begin{abstract}
\vspace{-0.2cm}
Despite rapid advances in face recognition, there remains a clear gap between the performance of still image-based face recognition and video-based face recognition, due to the vast difference in visual quality between the domains and the difficulty of curating diverse large-scale video datasets. This paper addresses both of those challenges, through an image to video feature-level domain adaptation approach, to learn discriminative video frame representations. The framework utilizes large-scale \textbf{unlabeled} video data to reduce the gap between different domains while transferring discriminative knowledge from large-scale labeled still images. Given a face recognition network that is pretrained in the image domain, the adaptation is achieved by (i) distilling knowledge from the network to a video adaptation network through feature matching, (ii) performing feature restoration through synthetic data augmentation and (iii) learning a domain-invariant feature through a domain adversarial discriminator. We further improve performance through a discriminator-guided feature fusion that boosts high-quality frames while eliminating those degraded by video domain-specific factors. Experiments on the YouTube Faces and IJB-A datasets demonstrate that each module contributes to our feature-level domain adaptation framework and substantially improves video face recognition performance to achieve state-of-the-art accuracy. We demonstrate qualitatively that the network learns to suppress diverse artifacts in videos such as pose, illumination or occlusion without being explicitly trained for them.
\end{abstract}

%%%%%%%%% INTRO
\vspace{-0.2in}
\section{Introduction}\label{sec:intro}
\vspace{-0.07in}

\begin{figure}[t]
\centering{\includegraphics[width = .95\linewidth]{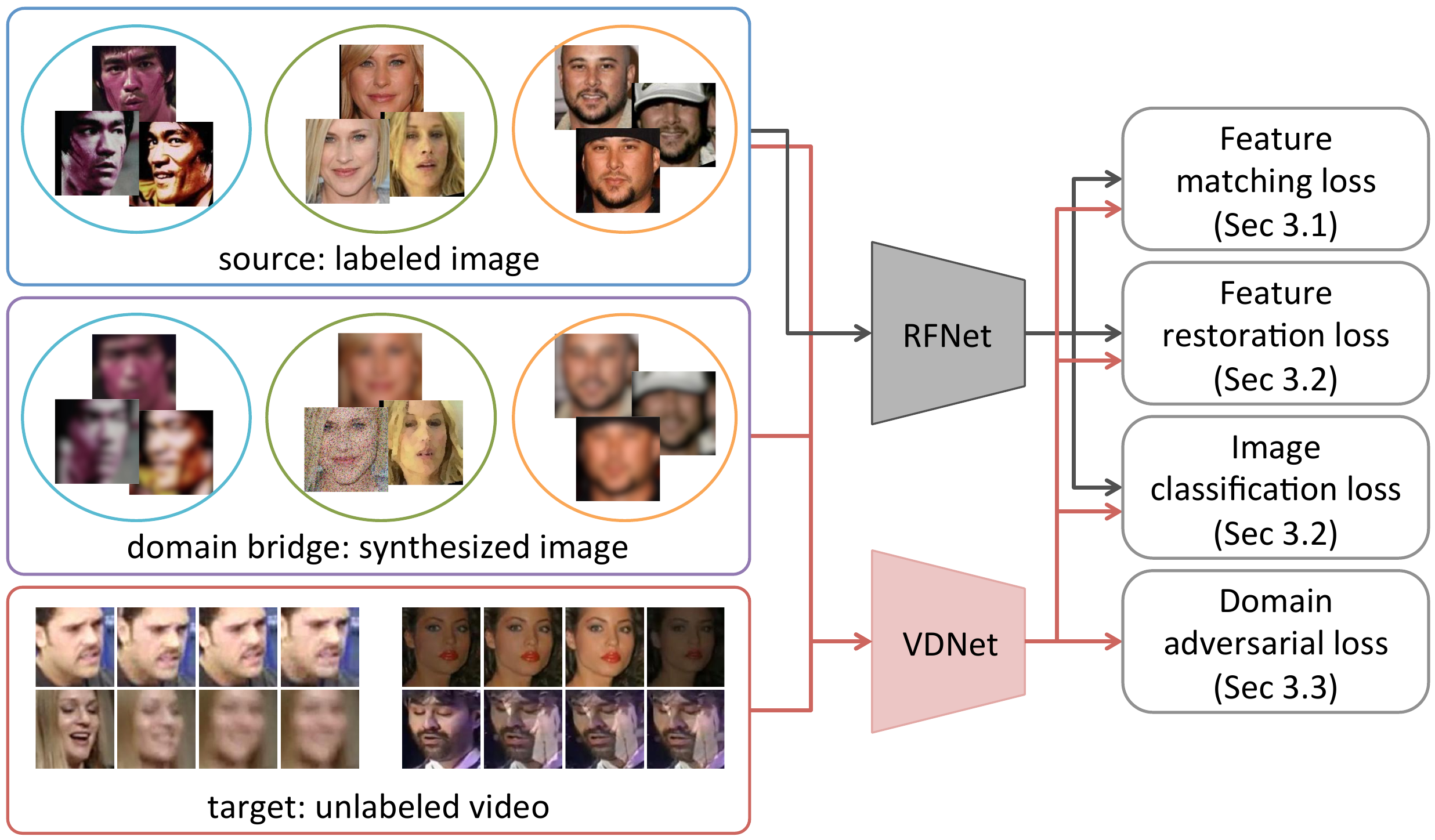}}
\caption{We propose an unsupervised domain adaptation method for video face recognition using large-scale unlabeled videos and labeled still images. To help bridge the gap between two domains, we introduce a new domain of synthesized images by applying a set of image transformations specific to videos such as motion blur to labeled images that simulates a video frame from still image. We utilize images, synthesized images, and unlabeled videos for domain adversarial training. Finally, we train a video domain-adapted network (VDNet) with domain adversarial loss (Section~\ref{sec:adversary}) as well as by distilling knowledge from pretrained reference network (RFNet) through feature matching (Section~\ref{sec:fm}), feature restoration and image classification (Section~\ref{sec:restoration}) losses.}
\label{fig:intro}
\vspace{-0.2in}
\end{figure}

Motion of objects or the observer in a video sequence is a powerful cue for perceptual tasks such as shape determination or identity recognition \cite{Cutting_1986,Gibson_1979}. For face recognition in computer vision, recent years have seen the success of emerging approaches in video face analysis \cite{vface-wolf2011,vface-li2013,vface-cui2013,vface-Wolf2013,vface-li2014} and several image-based face recognition engines \cite{deepface2014,vggface15} on video face recognition benchmarks \cite{vface-wolf2011,IJB}. But it is arguably true that these efforts have been outpaced by image-based face recognition engines that perform comparably or even better than human perception in certain settings \cite{deepface2014,sun2014deep,schroff2015facenet,LFWTech}. For example, a verification accuracy of $95.12\%$ is reported by \cite{schroff2015facenet} on the YouTube Faces dataset, much lower than $99.63\%$ on the LFW dataset.

Besides better understanding of training convolutional neural networks (CNNs), a key ingredient to the success of image-based face recognition is the availability of large-scale datasets of labeled face images collected from the web \cite{YD2014}. Thus, one source of difficulty for video face recognition is attributable to the lack of similar large-scale labeled datasets. The number of images used to train the state-of-the-art face recognition engines varies from $200$K~\cite{sun2014deep} to $200$M~\cite{schroff2015facenet} collected with at least $10$K different identities or as many as $8$M. In contrast, large-scale labeled video database is publicly available to date, such as YouTube face dataset (YTF)~\cite{vface-wolf2011}, only contains $3.4$K videos in total from $1.5$K different subjects. Although more frames may be labeled, it is difficult to collect a dataset with as many variations without a surge in dataset size and labeling effort.

An avenue for overcoming the lack of labeled training data in the video domain is to transform labeled still face images so that they look like images captured from videos. Video frames are likely to be degraded for multiple reasons such as motion or out-of-focus blur, compression noise or scale variations. Approaches such as \cite{vface-trunk2016} augment image-based training data with synthetic blur kernels and noise, to demonstrate moderate improvement in video face recognition. However, attempting to bridge the domain gap between images and videos with such an approach faces fundamental challenges -- first, it is non-trivial to sufficiently enumerate all types of blur kernels that degrade visual quality in videos, and second, it is not possible to model the transformation from images to videos with sufficient accuracy.

In this work, we propose a data-driven method for image to video domain adaptation for video face recognition. Instead of collecting a labeled video face dataset, we utilize large-scale unlabeled video data to reduce the gap between video and image domains, while retaining the discriminative power of large-scale labeled still images. To take advantage of labeled image data, Section \ref{sec:fm} proposes to transfer discriminative knowledge by distilling the distance metric through feature matching, from a reference network (RFNet) trained on a web-face dataset~\cite{YD2014} to our video face network (VDNet). A further avenue to leverage image domain labels is through the domain-specific data augmentation of Section \ref{sec:restoration}, whereby we degrade still images using synthetic motion blur, resolution variation, or video compression noise. Then, we train VDNet to be able to \emph{restore} the original representation of an image extracted from RFNet.

While the above augmentation is useful, its effectiveness is limited by the fact that types of artifacts in videos are too diverse to be enumerated. In Section \ref{sec:adversary}, we further regularize VDNet to reduce the domain gap by introducing a discriminator that learns to distinguish different domains, without any supervision such as identity labels or instance-level correspondence. Once trained, the score output by this discriminator is a measure of the confidence in the similarity of the feature representation of a video frame to that of a still image. This is a useful ability, since poor performance in video face recognition can often be attributed to some frames in a sequence that are of substantially poor quality. Consequently, Section \ref{sec:fuse} proposes a discriminator-guided weighted feature fusion to aggregate frames in each video, by assigning higher weights to ``image-like'' frames, that potentially have better quality among the others. Figure~\ref{fig:intro} illustrates our proposed framework.

In Section \ref{sec:exp}, we extensively evaluate the proposed framework on the YouTube Faces (YTF) dataset to demonstrate performance that surpasses prior state-of-the-art. We present ablation studies that demonstrate the importance of each of the above components. Interestingly, degradation factors such as blur, illumination or occlusions, automatically emerge in qualitative visualizations of frames within a sequence ranked by domain discriminator scores. 

The main contributions of this work are:
\begin{tight_itemize}
\vspace{-0.2cm}
\item We present a novel unsupervised domain adaptation algorithm from images to videos for face recognition in unlabeled videos.
\item We develop a feature-level domain adaptation to learn VDNet by distilling discriminative knowledge from pretrained RFNet through feature matching.
\item We propose a domain adversarial learning method that modulates the VDNet to learn a domain-invariant feature without needing to enumerate all causes of domain gap.
\item We design a method to train with synthetic data augmentation for feature-level restoration and to help the discriminator to discover domain differences.
\item We use the confidence score of the discriminator to develop an unsupervised feature fusion method that suppresses low quality frames.
\item We demonstrate the superiority of VDNet over existing methods with extensive experiments on YTF dataset, achieving state-of-the-art verification accuracy. We also demonstrate performance gains over baseline methods on the IJB-A dataset without supervised fine-tuning.
\end{tight_itemize}

\begin{figure*}[htbp]
\vspace{-0.1in}
\centering
{\includegraphics[width = .9\linewidth]{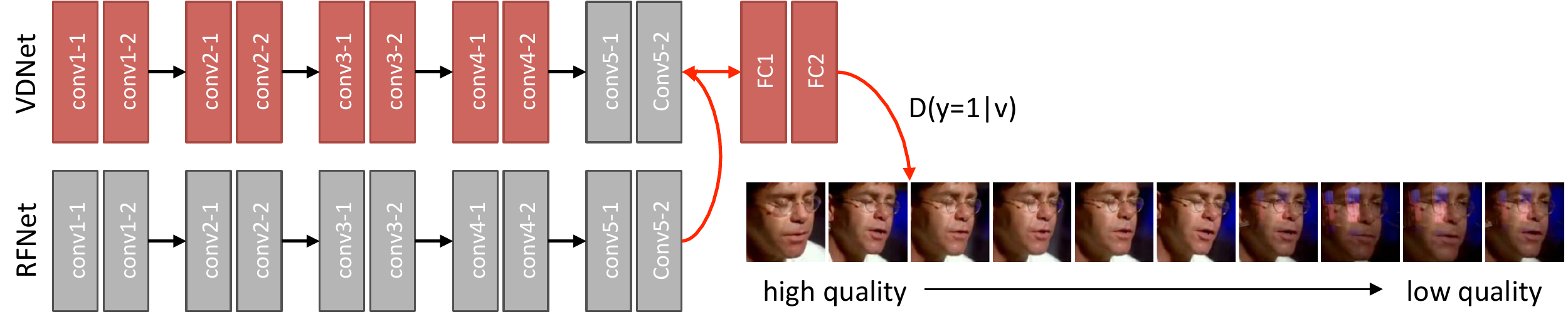}}
\caption{An illustration of network architecture for RFNet, VDNet and discriminator ($\mathcal{D}$). The red and gray blocks denote the trainable and fixed modules, respectively. VDNet not only shares the network architecture with RFNet, but also is initialized with the same network parameters. Once trained, $\mathcal{D}$ can sort the frames in a video sequence by indicating whether a frame is similar to images compatible to a face recognition engine and rejects those frames that are extremely ill-suited for face recognition.}
\label{fig:network}
\vspace{-4mm}
\end{figure*}

%%%%%%%%%% RELATED WORK
\vspace{-0.1in}
\section{Related Work}
\label{sec:related}
\vspace{-0.07in}
Our work falls into the class of problems on unsupervised domain adaptation~\cite{uda-long2013,uda-fernando2015,uda-TzengHZSD14,ganin2015unsupervised} that concerns adapting a classifier trained on a source domain (e.g., web images) to a target domain (e.g., video) where there is no labeled training data for target domain to fine-tune the classifier. Among those, feature space alignment and domain adversarial learning methods are closely related to our approach.

The basic idea of feature space alignment is to minimize the distance between domains in the feature space through learning a transformation of source to target features \cite{uda-fernando2015,uda-saenko2010,uda-fernando2013,Wang_Transformation,zhou2016view}, or a joint adaptation layer that embeds features into a new domain-invariant space \cite{uda-long2013,uda-TzengHZSD14}. Specifically, Tzeng et al.~\cite{uda-TzengHZSD14} use two CNNs for source and target domain with shared weights and the network is optimized for classification loss in the source domain as well as domain difference measured by the maximum mean discrepancy (MMD) metric. Gupta et al.~\cite{gupta2016cross} consider a similar network architecture for cross-modality supervision transfer.

There also exists a body of work on unsupervised domain adaptation and transfer with adversarial learning \cite{goodfellow2014generative}, where the domain difference is measured by a discriminator network $\mathcal{D}$ \cite{yoo2016pixel,liu2016coupled,gan-TaigmanPW16}. For example, \cite{yoo2016pixel,gan-TaigmanPW16} consider cross-domain transfer of images from one style to another without instance-level correspondence between domains using adversarial loss. Coupled GAN~\cite{liu2016coupled} constructs individual networks for each domain with partially shared higher-layer parameters for generator and discriminator to generate coherent images of two domains. Unlike the above works that generate images in the target domain, we consider feature-level domain adaptation.

For feature-level domain adaptation using adversarial learning, domain adversarial neural network (DANN)~\cite{ganin2015unsupervised} appends domain classifier to high-level features and introduces a gradient reversal layer for end-to-end learning via backpropagation while avoiding cumbersome minimax optimization of adversarial training. The goal of DANN is to transfer discriminative classifier from source to target domain, which implicitly assumes the label spaces of two domains are equivalent (or at least the label space of target domain is the subset of that of source domain). Our work is to transfer discriminative \emph{distance metric} and hence there is no such restriction in label space definition. In addition, we propose domain-specific synthetic data augmentation to further enhance the performance of domain adaptation and use discriminator outputs for feature fusion.

%%%%%%%%%%% METHOD
\vspace{-0.04in}
\section{Domain Adaptation from Image to Video}
\label{sec:DAnet}
\vspace{-0.07in}

As previewed in Section \ref{sec:intro}, curating large-scale video datasets with identity labels is an onerous task, but there do exist such datasets for still images. This makes it natural to consider image to video domain adaptation. However, the representation gap is a challenging one to bridge due to blur, compression, motions and other artifacts in videos. This section tackles the challenge by introducing a set of domain adaptation objectives that allow our video face recognition network (VDNet) to be trained on large-scale unlabeled videos in $\mathcal{V}$, while taking advantage of supervised information from labeled web-face images in $\mathcal{I}$.

\vspace{-1mm}
\subsection{Distilling Knowledge by Feature Matching}
\vspace{-1mm}		 
\label{sec:fm}
To take advantage of labeled web-face images, we train VDNet by distilling discriminative knowledge from a face recognition engine pretrained on a labeled web-face dataset, which we call a reference network (RFNet). Unlike previous works that distill knowledge through class probabilities~\cite{uda-distill}, we do so by matching feature representations between two networks, since we do not have access to labeled videos. Let $\phi(\cdot):\mathbb{R}^{D}\rightarrow\mathbb{R}^{K}$ be a feature generation operator of VDNet and $\psi(\cdot):\mathbb{R}^{D}\rightarrow\mathbb{R}^{K}$ be that of RFNet. The feature matching (FM) loss is defined on an image $x\in\mathcal{I}$ as:
\begin{equation}
\mathcal{L}_{\text{FM}} = \frac{1}{|\mathcal{I}|}\sum_{x\in\mathcal{I}}\Vert \phi(x) - \psi(x) \Vert_{2}^{2}
\end{equation}
The FM loss allows VDNet to maintain a certain degree of discriminative information for face identity recognition. With regards to network structure, VDNet can be very flexible as long as the matching feature has the same dimensionality with that of RFNet. In practice, we use the same network architecture between VDNet and RFNet. Moreover, we initialize the network parameters of VDNet with RFNet and freeze network parameters for a few higher layers to further maintain discriminative information learned from labeled web-face images, as illustrated in Figure \ref{fig:network}. Note that while more complex distillation methods and architectures are certainly possible, our intent is simply a strong initialization for VDNet, for which these choices suffice.

\vspace{-0.04in}	
\subsection{Adaptation via Synthetic Data Augmentation}
\vspace{-0.05in}	
\label{sec:restoration}
Data augmentation has been widely used for training very deep CNNs with limited amount of training data as it prevents overfitting and enhances generalization ability. In addition to generic data transformations such as random cropping or horizontal flips, applying data transformation that is specific to the target domain has been shown to be effective~\cite{vface-trunk2016}. To generalize to video frames, we consider data augmentation by applying transformations such as linear motion blur, image resolution (scale) variation or video compression noise, which are the most typical causes of quality degradation in video. We train VDNet to ``restore'' the original RFNet representation of an image without data augmentation through the feature restoration (FR) loss:
\vspace{-0.2cm}
\begin{equation}
\mathcal{L}_{\text{FR}} = \frac{1}{|\mathcal{I}|}\sum_{x\in\mathcal{I}} \mathbb{E}_{B(\cdot)}\Big[\Vert\phi(B(x)) - \psi(x) \Vert_{2}^2\Big]
\vspace{-0.2cm}
\end{equation}
where $B(\cdot):\mathbb{R}^{D}\rightarrow \mathbb{R}^{D}$ is an image transformation kernel and $\mathbb{E}_{B(\cdot)}$ is the expectation over the distribution of $B(\cdot)$. In this work, we consider three types of image transformations with the following parameters:
\begin{itemize}[leftmargin=10pt,itemsep=10pt,nolistsep]
	\item Linear motion blur: kernel length is randomly selected in $\left(5,15 \right)$ and kernel angle is selected in $\left( 10,30\right)$.
	\item Scale variation: we rescale an image as small as $\frac{1}{6}$ of the original image size.
	\item JPEG compression: the quality parameter is set randomly in $\left( 30, 75\right)$.
\end{itemize}
These augmentations are applied in sequence to an image with probability of $0.5$ for each noise process.

Taking advantage of labeled training examples from image domain, one can also use standard metric learning objectives to learn discriminative metric that generalizes to low-quality images defined by aforementioned blur kernels.
We adopt N-pair loss~\cite{NIPS2016_6200}, which is shown to be effective at learning deep distance metric from large number of classes.
Given N pairs of examples from N different classes $\{(x_{i}, x_{i}^{+})\}_{i=1}^{N}$ with individual synthetic data augmentation $B_{i}(\cdot)$, the N-pair loss is defined as follows: 
\begin{equation}
\mathcal{L}_{\text{IC}} = -\frac{1}{N}\sum_{i=1}^{N}\log\frac{\exp(\phi(B_{i}(x_{i}^{+}))^{\top}\psi(x_{i}))}{\sum_{n=1}^{N}\exp(\phi(B_{i}(x_{i}^{+}))^{\top}\psi(x_{n}))}
\end{equation}
We note that N-pair loss could be one example of an objective function for metric learning with synthetic augmentation, but can be replaced with other standard metric learning objectives such as contrastive loss~\cite{chopra2005learning} or triplet loss~\cite{schroff2015facenet}. 

\vspace{-0.04in}			 
\subsection{Adaptation via Domain Adversarial Learning}
\label{sec:adversary}
\vspace{-0.05in}	
Although data augmentation has been successful in many computer vision applications, the types of transformation between source and target domains are not always known, that is, there might be many unknown factors of variation between two domains. Moreover, modeling such transformations is challenging even if they are known, so we may need to resort to an approximation in many cases. Therefore, it is difficult to close the gap between two domains. Rather than attempting to exhaustively enumerate or approximate different types of transformations between two domains, we learn them from large-scale unlabeled data and facilitate the recognition engine to be robust to those transformations. 

Adversarial learning~\cite{goodfellow2014generative} provides a good framework to approach the above problem, whereby the generator, that is, VDNet, is regularized to close the gap between two domains, where the domain difference is captured by the discriminator.
The adversarial loss with two domains $\mathcal{I}$ and $\mathcal{V}$ is defined over the expectation of all training samples:
%can be written as follows:
%
\begin{align}
\mathcal{L}_{\text{D}}  = & - \mathbb{E}_{x\in\mathcal{I}}\big[ \log \mathcal{D}(y=1|\phi(x))\big]\\\nonumber
&- \mathbb{E}_{x\in\mathcal{V}}\big[\log \mathcal{D}(y=2|\phi(x))\big] \\
\mathcal{L}_{\text{Adv}} = & -\mathbb{E}_{x\in\mathcal{V}}\big[\log \mathcal{D}(y=1|\phi(x))\big] 
\end{align}
The discriminator ($\mathcal{D}$) is defined on top of VDNet that already induces highly abstract features from a deep CNN. Thus, the architecture of $\mathcal{D}$ can be very simple, such as two or three fully-connected layer networks. Unlike several recent applications of adversarial frameworks on image translation~\cite{Ledig_2017_CVPR,taigman2016unsupervised}, $\mathcal{D}$ is not distinguishing between generated and real images in pixel space, rather between feature representations. We argue that this is desirable due to the relative maturity of feature learning for face recognition as opposed to high-quality image generation.

Note that adversarial loss allows utilizing a large volume of unlabeled video data to train VDNet without additional labeling effort. However, the loss can only match representations between two domains in a global manner and the effect would be marginal if the contrast between two domains is small or the discriminator cannot distinguish them well. 
As a result, we can still take advantage of synthetic data augmentation to guide the discriminator, either to realize the difference between domains or to discriminate additional domain differences from known synthetic transformations.
This naturally leads us to two different discriminator types, one with two-way classifier between image ($\mathcal{I}$) and synthesized image and video ($B(\mathcal{I})\cup\mathcal{V}$) or the other with a three-way classifier among image, synthesized image, and video. 

\vspace{-0.3cm}
\paragraph{Two-way $\mathcal{D}$.} We use a two-way softmax classifier as $\mathcal{D}$ to discriminate between the image domain ($y=1$) and the domain of synthesized images and videos ($y=2$). While the original images are from the image domain, both synthetically degraded images as well as random video frames are trained to belong to the same domain as follows:
\begin{align}
\mathcal{L}_{\mathcal{D}}  = & - \mathbb{E}_{x\in\mathcal{I}}\big[\log \mathcal{D}(y=1|\phi(x))\big]\nonumber\\
& - \mathbb{E}_{x\in B(\mathcal{I})\cup\mathcal{V}}\big[\log \mathcal{D}(y=2|\phi(x))\big]\\
\mathcal{L}_{\text{Adv}} = & -\mathbb{E}_{x\in B(\mathcal{I})\cup\mathcal{V}}\big[ \log \mathcal{D}(y=1|\phi(x))\big]
\end{align}
Since the contrast between two classes becomes apparent by including synthetic images for the second class, the transformations in the video domain that are similar to synthetic image transformations can be easily restored.

\vspace{-0.3cm}
\paragraph{Three-way $\mathcal{D}$.} We use a three-way softmax classifier as $\mathcal{D}$ to discriminate images ($y=1$), synthesized images ($y=2$) and video frames ($y=3$) into three different categories.
\begin{align}
\mathcal{L}_{\mathcal{D}}  = & - \mathbb{E}_{x\in\mathcal{I}}\big[ \log \mathcal{D}(y=1|\phi(x))\big]\nonumber\\
& - \mathbb{E}_{x\in\mathcal{B(I)}}\big[\log \mathcal{D}(y=2|\phi(x))\big]\\
& - \mathbb{E}_{x\in\mathcal{V}}\big[\log \mathcal{D}(y=3|\phi(x))\big] \nonumber\\
\mathcal{L}_{\text{Adv}} = & -\mathbb{E}_{x\in B(\mathcal{I})\cup\mathcal{V}}\big[ \log \mathcal{D}(y=1|\phi(x))\big] 
\end{align}
Unlike the two-way network, the three-way network aims to distinguish video frames from not only the image domain but also synthetically degraded images. Therefore, it may not learn a VDNet with as strong restoration capability to synthetic transformations as with two-way discriminator, but aims to find additional factors of variation between image or synthetic image and video domains.

Overall, the objective function is written as follows:
\begin{equation}
\mathcal{L} = \mathcal{L}_{\text{FM}} + \alpha\mathcal{L}_{\text{FR}} + \beta\mathcal{L}_{\text{IC}} + \gamma\mathcal{L}_{\text{Adv}}\label{eq:objective-all}
\end{equation}

\vspace{-0.07in}	
\subsection{Discriminator-Guided Feature Fusion}
\vspace{-0.05in}		 
\label{sec:fuse}
As noted by Yang et al.~\cite{Yang_2017_CVPR}, the quality evaluation of each frame is important for video face recognition since not all frames contribute equally. Moreover, it is important to discount frames that are extremely noisy due to motion blur or other noise factors, in favor of those that are better for recognition. Our discriminator is already trained to distinguish still images from blurred ones or video frames, so its output may already be used as a confidence score at for each frame being a high quality image.
Training with the domain contrast between image, blurred image and video, the discriminator is ready to provide a confidence score at test time, for each frame being a ``high-quality web image'' ($\mathcal{D}(y=1|\phi(v))$). 
Specifically, with the confidence score from the discriminator, the aggregated feature vector for a video $V$ with frames $v$ is represented as a weighted average of feature vectors as follows:
\begin{equation}
\phi_{V} = \frac{\sum_{v\in V} \mathcal{D}\left(y=1|\phi\left( v\right) \right) \cdot\phi\left(v\right) }{\sum_{v\in V}\mathcal{D}\left(y=1|\phi\left( v\right) \right)}.
\label{eq:fuse}
\end{equation}
Note that this target domain of web images comes with large-scaled labeled training examples to train a discriminative face recognition engine. Thus, the discriminator serves a dual role of guiding both the feature-level domain adaptation and a fusion weighted by confidence in the fitness of a frame for a face recognition engine.

%%%%%%%%%% DETAILS

\vspace{-0.05in}	
\section{Implementation Details}
\label{sec:implementation}
\vspace{-0.05in}	
We provide detailed information of network architectures for the RFNet, the VDNet, and the discriminator.

\vspace{-0.05in}	
\subsection{Face Recognition Engine}			 
\label{sec:fr-npair}
\vspace{-0.05in}	
Our face recognition engine is also on a deep CNN trained on CASIA-webface dataset~\cite{YD2014}. The network architecture is similar to the ones used in~\cite{YD2014,NIPS2016_6200}, which contains 10 layers of $3\times3$ convolution followed by ReLU nonlinearities with 4 max pooling layers with stride 2 and one average pooling layer with stride 7, except for that our network uses strided convolution to replace max pooling and maxout units~\cite{goodfellow2013maxout} for every other convolution layer instead of ReLU layers. Please see supplementary material for more details. The model is trained with the deep metric learning objective called N-pair loss~\cite{NIPS2016_6200} as described in Section~\ref{sec:restoration}. Our implementation is based on Torch~\cite{collobert:2011c} and $N=1080$ (N-pair loss pushes (N-1) negative examples at the same time while pulling a single positive example) is used on 8 GPUs for training. Faces are detected and aligned using keypoints~\cite{yu2016deep} and $100\times100$ gray-scale image patches randomly cropped from $110\times110$ resized face images are fed to network for training. The model achieves $98.85\%$ verification accuracy on the Labeled Faces in the Wild dataset~\cite{LFWTech}. 

The RFNet is the same as face recognition engine and the parameters are fixed. The VDNet is initialized the same as RFNet but the parameters are updated for all layers except for the last two convolution layers, as illustrated in Figure~\ref{fig:network}.

\vspace{-0.05in}	
\subsection{Discriminator}
\vspace{-0.05in}	
We apply a similar network architecture of $\mathcal{D}$ for two and three-way discriminators. For the discriminator $\mathcal{D}$, we use a simple neural network with two fully-connected layers ($320-160-\text{ReLU}-3$) as shown in Figure~\ref{fig:network}. For two-way networks, we replace the output channel of last fully-connected layer from three to two. We train the network using Adam optimizer~\cite{adam} with $\beta_{1} = 0.9$, $\beta_2 = 0.999$, and learning rate of $0.0003$, while setting $\alpha=\beta=\gamma=1$. Details of our network architecture and hyper parameters can be found in supplementary material.

\begin{table*}[t]
\centering
\caption{Video face recognition accuracy and standard error on the YTF dataset. Image-classification loss (IC), feature matching loss (FM), feature restoration loss (FR) and adversarial loss (Adv) are applied for training. For feature restoration loss, we consider three types of data augmentation, namely, linear motion blur (M), scale variation (S), or JPEG compression noise (C). The best performer and those with overlapping standard error are boldfaced.\label{tab:ytf_summary}}
\vspace{0.04in}
\footnotesize
\begin{tabular}{c|c|c|c|c|c||c|c|c|c|c}
\hline
Model & IC & FM & FR & Adv & fusion & 1 (fr/vid) & 5 (fr/vid) & 20 (fr/vid) & 50 (fr/vid) & all\\
\hline
\multirow{2}{*}{baseline} & \multicolumn{4}{c|}{\multirow{2}{*}{--}} & -- & 91.12{\footnotesize$\pm 0.318$} & 93.17{\footnotesize$\pm 0.371$} & 93.62{\footnotesize$\pm 0.430$} & 93.74{\footnotesize$\pm 0.443$} & 93.78{\footnotesize$\pm 0.498$}\\
& \multicolumn{4}{c|}{} & \checkmark & -- & 93.30{\footnotesize$\pm 0.362$} & 93.72{\footnotesize$\pm 0.428$} & 93.80{\footnotesize$\pm 0.444$} & 93.94{\footnotesize$\pm 0.493$}\\
\hline
A & \checkmark & -- & M/S & -- & -- & 91.37{\footnotesize$\pm 0.334$} & 92.97{\footnotesize$\pm 0.381$} & 93.42{\footnotesize$\pm 0.399$} & 93.43{\footnotesize$\pm 0.384$} & 
93.32{\footnotesize$\pm 0.443$}\\
\hline
B & \checkmark & \checkmark & M/S & -- & -- & 91.44{\footnotesize$\pm 0.348$} & 93.46{\footnotesize$\pm 0.392$} & 93.84{\footnotesize$\pm 0.433$} & 93.95{\footnotesize$\pm 0.443$} & 93.94{\footnotesize$\pm 0.507$}\\
\hline
C & \checkmark & \checkmark & M/S/C & -- & -- & 91.68{\footnotesize$\pm 0.320$} & 93.52{\footnotesize$\pm 0.323$} & 93.94{\footnotesize$\pm 0.337$} & 93.90{\footnotesize$\pm 0.361$} & 93.82{\footnotesize$\pm 0.383$}\\
\hline
D & \checkmark & \checkmark & -- & two-way & -- & 91.38{\footnotesize$\pm 0.350$} & 93.74{\footnotesize$\pm 0.354$} & 94.04{\footnotesize$\pm 0.375$} & 94.23{\footnotesize$\pm 0.379$} & 94.36{\footnotesize$\pm 0.346$}\\
\hline
\multirow{2}{*}{E} & \multirow{2}{*}{\checkmark} & \multirow{2}{*}{\checkmark} & \multirow{2}{*}{M/S/C} & \multirow{2}{*}{two-way} & -- & 92.39{\footnotesize$\pm 0.315$} & 94.72{\footnotesize$\pm 0.306$} & \bf{95.13}{\footnotesize$\pm 0.263$} & \bf{95.13}{\footnotesize$\pm 0.286$} & \bf{95.22}{\footnotesize$\pm 0.319$}\\
& & & & & \checkmark & -- & 94.73{\footnotesize$\pm 0.270$} & \bf{95.14}{\footnotesize$\pm 0.229$} & \bf{95.13}{\footnotesize$\pm 0.261$} & \bf{95.16}{\footnotesize$\pm 0.284$}\\
\hline
\multirow{2}{*}{F} & \multirow{2}{*}{\checkmark} & \multirow{2}{*}{\checkmark} & \multirow{2}{*}{M/S/C} & \multirow{2}{*}{three-way} & -- & 92.17{\footnotesize$\pm 0.353$} & 94.44{\footnotesize$\pm 0.343$} & \bf{94.90}{\footnotesize$\pm 0.345$} & \bf{94.98}{\footnotesize$\pm 0.354$} & \bf{95.00}{\footnotesize$\pm 0.415$}\\
&  &  &  &  & \checkmark & -- & 94.52{\footnotesize$\pm 0.356$} & \bf{95.01}{\footnotesize$\pm 0.352$} & \bf{95.15}{\footnotesize$\pm 0.370$} & \bf{95.38}{\footnotesize$\pm 0.310$}\\
\hline
\end{tabular}
\vspace{-0.18in}
\end{table*}

\vspace{-0.05in}	
\section{Experimental Results}\label{sec:exp}
\vspace{-0.05in}	
We evaluate the performance of our proposed unsupervised domain adaptation framework, by first providing the baseline methods in Section~\ref{sec:base}, and then performing an ablation study for each component of the proposed approach on YouTube Faces (YTF) dataset~\cite{vface-wolf2011} in Section \ref{sec:ytb}. We further evaluate the model trained on the YTF dataset to IJB-A~\cite{IJB}, which demonstrates the generalization capabilities of the proposed approach, as presented in Section~\ref{sec:ijb}.

\vspace{-0.05in}
\subsection{Evalutation Protocol}\label{sec:base}
\vspace{-0.05in}
The standard application of image-based face recognition engine for video face recognition is to first apply the face recognition engine to each frame and then aggregate extracted features from individual frames to obtain a single representation of videos. For baselines, we follow the standard protocol of extracting L2 normalized features from each frame and its horizontally flipped image followed by temporal average pooling over frames per video. For discriminator-guided feature fusion, we follow Equation~\eqref{eq:fuse} to obtain a video representation from individual L2 normalized features. We compute inner product between two video representations for similarity metric.

\vspace{-1mm}		 
\subsection{YouTube Faces Dataset}\label{sec:ytb}
\vspace{-1mm}
The YTF dataset contains $3425$ videos of unconstrained face images from $1595$ different people with the average length of $181.3$ frames per video. Ten folds of video pairs are available for verification experiments, where each fold is composed of 250 positive and negative pairs with no overlapping identity between different folds. We use videos in 8 training folds in addition to CASIA-webface dataset to train a VDNet, but no identity label from video is used.

Six networks (A-F) with different combinations of feature matching (FM), feature restoration (FR) with various data augmentation methods, adversarial training (Adv) and discriminator-guided feature fusion are presented in Table~\ref{tab:ytf_summary}. 

\vspace{-0.4cm}
\paragraph{Feature matching loss.}
The FM loss enforces VDNet to learn similar representations as those produced by RFNet on labeled still images, which is one of the key contributors to a good initialization for our method. Combination of image classification (IC) and FM reduces to training the baseline model. The effectiveness of FM loss can be seen by comparing accuracies of model A and B in Table~\ref{tab:ytf_summary}. A significant performance drop is observed for the model trained without FM loss but only with the FR loss, when it is evaluated with more number of frames per video. We hypothesize that while feature restoration loss drives the VDNet to match the representation of low-quality images to their high-quality counterpart, the representation of the original high-quality images are severely damaged, causing the model to lose its superior performance on high-quality images. By requiring the network to work well on both high-quality as well as low-quality images with FM loss, we observe significant improvement when evaluated with larger number of frames per video (for example, from $93.32\%$ of model A to $93.94\%$ of model B with all frames per video).

\vspace{-0.4cm}
\paragraph{Feature restoration loss.}
We consider three types of data augmentation described in Section~\ref{sec:restoration}, with different combinations presented with model B and C in Table~\ref{tab:ytf_summary}. Overall, FR loss moderately improves accuracy compared to the baseline models. Specifically, we observe that compression noise is quite effective at feature-level restoration when used along with linear motion blur and scale variations.

When combined with adversarial loss, we observe more significant improvement with feature restoration loss. For example, model E and F reduce the relative error by $11.7\%$ and $9.2\%$ compared to model D, respectively, on single frame per video evaluation regime and $15.6\%$ and $13.0\%$ when using randomly selected 50 frames per video for evaluation.

\begin{table*}[t]
\centering
\caption{1:1 verification and rank-K identification accuracy and standard error on IJB-A dataset. The model F is compared to the baseline method described in Section~\ref{sec:fr-npair}. Ten image crops (4 corners + 1 center + horizontal flip) are evaluated and fused together with uniform weights or discriminator confidence score. We also evaluate after removing images with significant localization errors ($^*$). \label{tab:ijba_summary}}
\vspace{0.04in}
\footnotesize
\begin{tabular}{c|c||c|c|c||c|c|c}
\hline
\multirow{2}{*}{Model} & \multirow{2}{*}{fusion} & \multicolumn{3}{c||}{1:1 Verification TAR} & \multicolumn{3}{c}{1:N Identification Rank-K Accuracy} \\\cline{3-8}
& & FAR=0.001 & FAR=0.01 & FAR=0.1 & Rank-1 & Rank-5 & Rank-10\\
\hline
baseline & -- & 0.526{\footnotesize$\pm 0.031$} & 0.757{\footnotesize$\pm 0.025$} & 0.951{\footnotesize$\pm 0.006$} & 0.855{\footnotesize$\pm 0.014$} & 0.950{\footnotesize$\pm 0.013$} & 0.970{\footnotesize$\pm 0.007$} \\
baseline & -- & 0.539{\footnotesize$\pm 0.013$} & 0.773{\footnotesize$\pm 0.008$} & 0.954{\footnotesize$\pm 0.002$} & 0.864{\footnotesize$\pm 0.004$} & 0.951{\footnotesize$\pm 0.003$} & 0.970{\footnotesize$\pm 0.002$} \\
baseline$^{*}$ & -- & 0.646{\footnotesize$\pm 0.012$} & 0.846{\footnotesize$\pm 0.005$} & 0.968{\footnotesize$\pm 0.001$} & 0.902{\footnotesize$\pm 0.003$} & 0.959{\footnotesize$\pm 0.003$} & 0.971{\footnotesize$\pm 0.001$} \\
\hline
F & -- & 0.531{\footnotesize$\pm 0.016$} & 0.800{\footnotesize$\pm 0.008$} & 0.963{\footnotesize$\pm 0.002$} & 0.869{\footnotesize$\pm 0.003$} & 0.954{\footnotesize$\pm 0.003$} & 0.970{\footnotesize$\pm 0.002$} \\
F & \checkmark & 0.584{\footnotesize$\pm 0.018$} & 0.828{\footnotesize$\pm 0.008$} & 0.962{\footnotesize$\pm 0.001$} & 0.879{\footnotesize$\pm 0.004$} & 0.955{\footnotesize$\pm 0.003$} & 0.970{\footnotesize$\pm 0.002$} \\
F$^{*}$ & \checkmark & 0.649{\footnotesize$\pm 0.022$} & 0.864{\footnotesize$\pm 0.007$} & 0.970{\footnotesize$\pm 0.001$} & 0.895{\footnotesize$\pm 0.003$} & 0.957{\footnotesize$\pm 0.002$} & 0.968{\footnotesize$\pm 0.002$}\\
\hline
Wang et al.~\cite{wang2015face} & -- & 0.510{\footnotesize$\pm 0.019$} & 0.729{\footnotesize$\pm 0.011$} & 0.893{\footnotesize$\pm 0.004$} & 0.822{\footnotesize$\pm 0.007$} & 0.931{\footnotesize$\pm 0.004$} & --\\
DCNN$_{\text{all}}$~\cite{chen2016unconstrained} & -- & -- & 0.787{\footnotesize$\pm 0.014$} & 0.947{\footnotesize$\pm 0.003$} & 0.860{\footnotesize$\pm 0.007$} & 0.943{\footnotesize$\pm 0.005$} & --\\
Yang et al.~\cite{Yang_2017_CVPR} & Mean L2 & 0.688{\footnotesize$\pm 0.025$} & 0.895{\footnotesize$\pm 0.005$} & 0.978{\footnotesize$\pm 0.001$} & 0.916{\footnotesize$\pm 0.004$} & 0.973{\footnotesize$\pm 0.002$} & 0.980{\footnotesize$\pm 0.001$}\\
Yang et al.~\cite{Yang_2017_CVPR} & NAN & 0.860{\footnotesize$\pm 0.004$} & 0.933{\footnotesize$\pm 0.003$} & 0.979{\footnotesize$\pm 0.001$} & 0.954{\footnotesize$\pm 0.002$} & 0.978{\footnotesize$\pm 0.001$} & 0.984{\footnotesize$\pm 0.001$}\\
\hline
\end{tabular}
\vspace{-0.2in}
\end{table*}

\vspace{-0.4cm}
\paragraph{Domain adversarial loss.}
In addition to feature restoration loss through synthetic data augmentation, domain adversarial training between high-quality web images and the videos contributes to reducing the gap between two domains. To demonstrate its effectiveness, we train the model only with Adv loss with random video frames as the additional input source with the ``fake'' labels (while random face images associate to the ``real'' ones), and compared to the baseline model. As shown in Table~\ref{tab:ytf_summary}, model D consistently outperforms the baseline one with different number of random frames per video. Note that the ``two-way'' in model D denotes a binary classification between random face images and video frames, without any artificially degraded sample.

\begin{figure*}[t!]
\centering
\begin{tabular}{@{\hspace{0mm}}c@{\hspace{0mm}}c}
\centering
\includegraphics[width=0.46\textwidth]{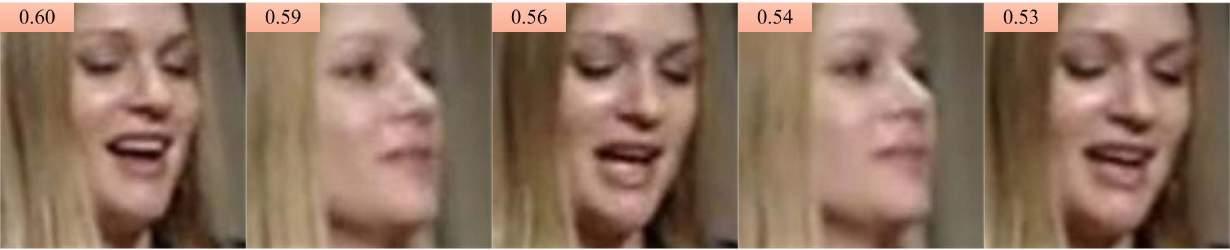}&\quad
\includegraphics[width=0.46\textwidth]{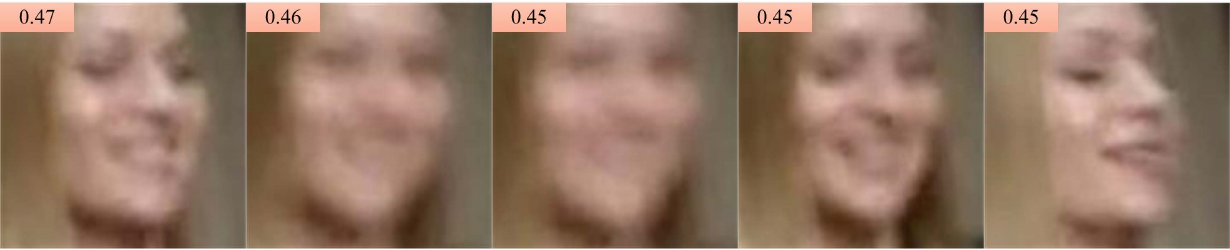}\\
\includegraphics[width=0.46\textwidth]{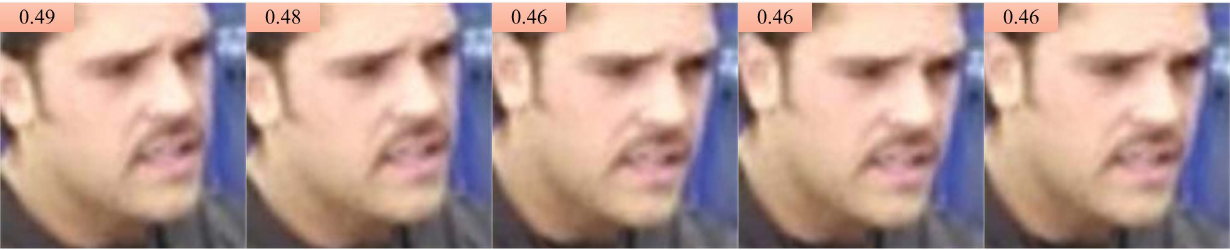}&\quad
\includegraphics[width=0.46\textwidth]{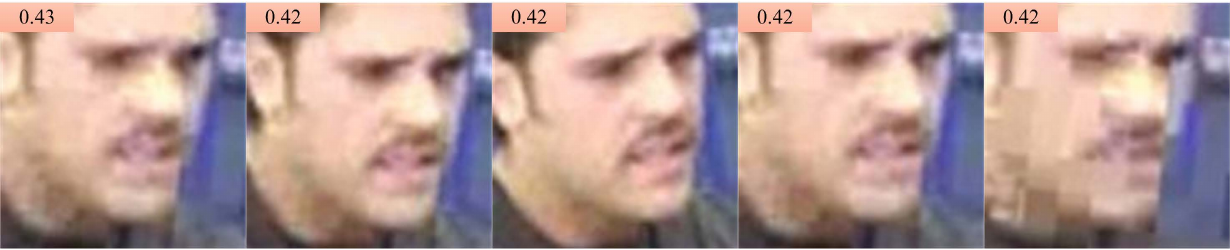}\\
\includegraphics[width=0.46\textwidth]{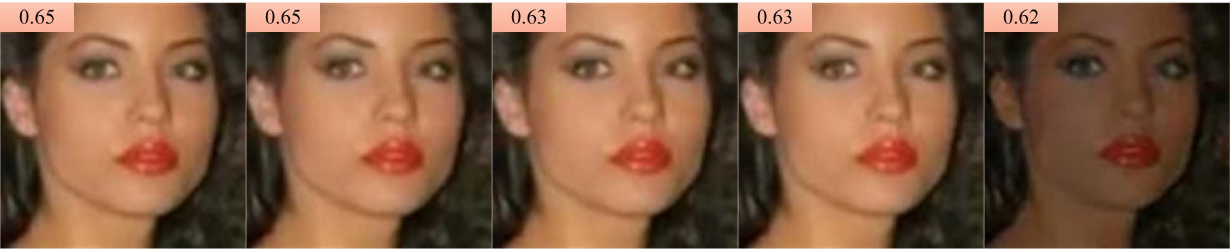}&\quad
\includegraphics[width=0.46\textwidth]{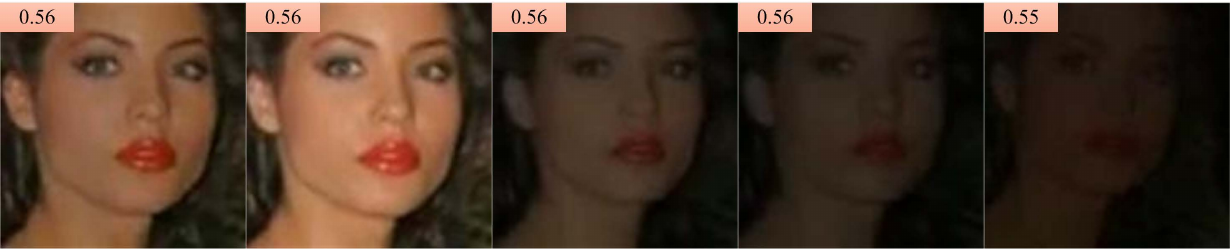}\\
\includegraphics[width=0.46\textwidth]{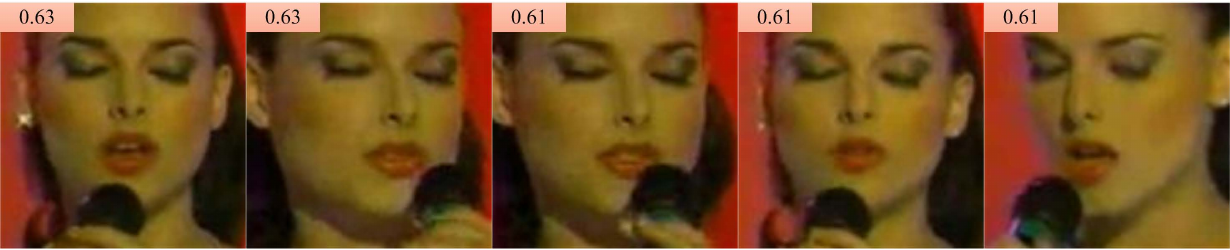}&\quad
\includegraphics[width=0.46\textwidth]{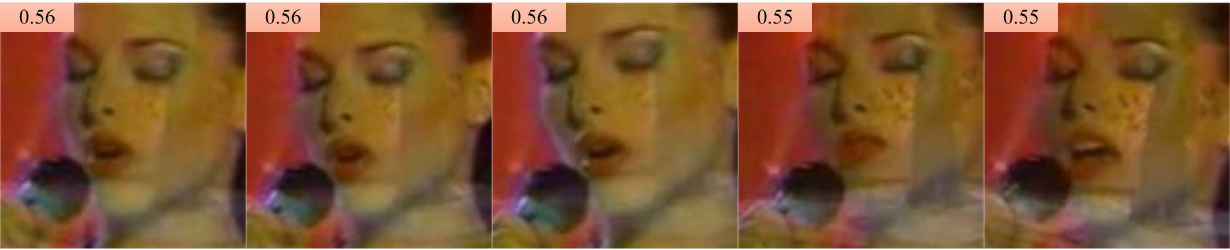}\\
\includegraphics[width=0.46\textwidth]{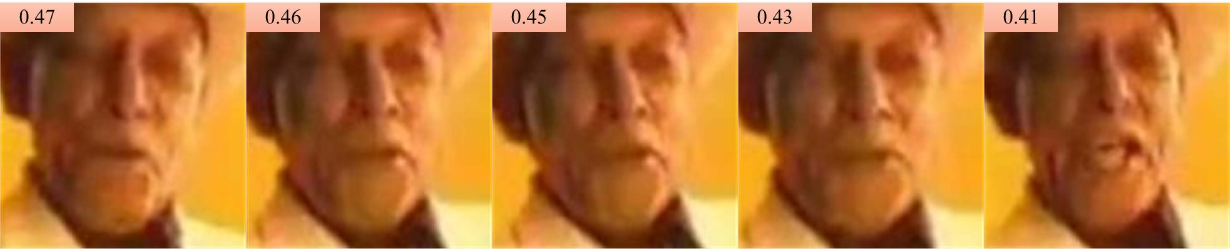}&\quad
\includegraphics[width=0.46\textwidth]{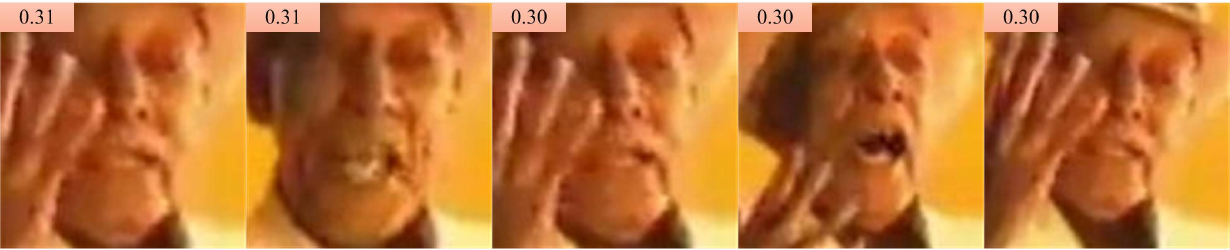}\\
\includegraphics[width=0.46\textwidth]{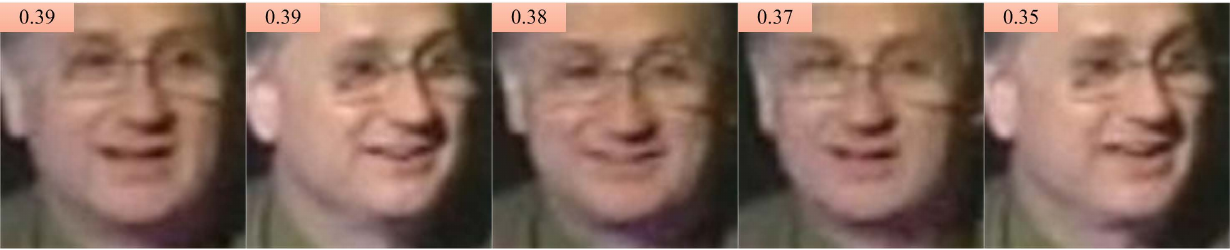}&\quad
\includegraphics[width=0.46\textwidth]{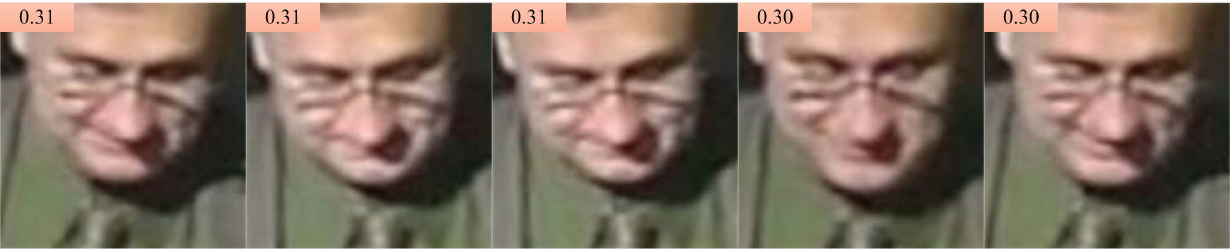}\\
\includegraphics[width=0.46\textwidth]{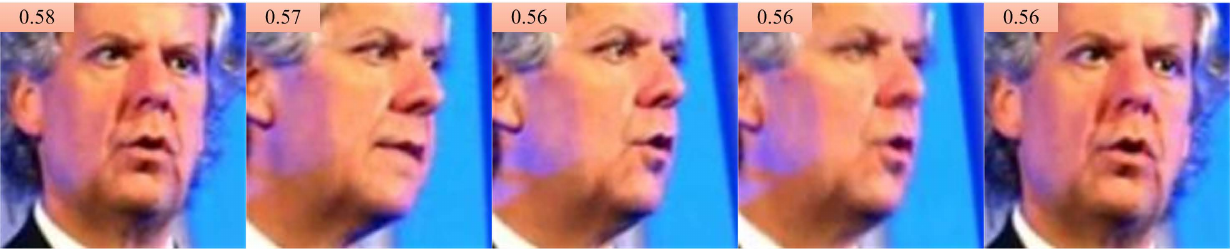}&\quad
\includegraphics[width=0.46\textwidth]{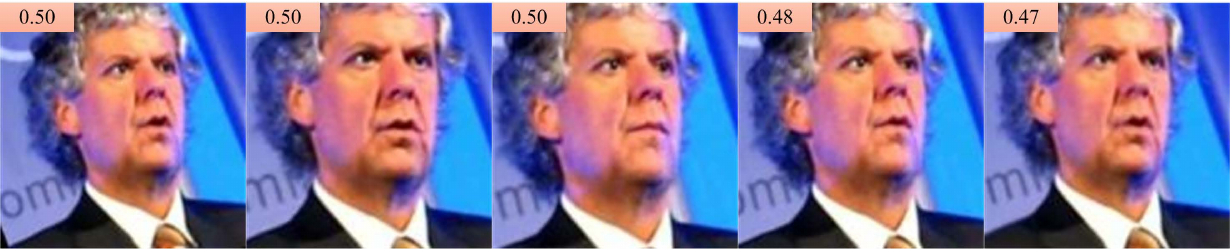}\\
{ (a) top-5}& {(b) bottom-5 } \\
\end{tabular}
\vspace{0.04in}
\caption{We sort the frames within a sequence in a descending order with respect to the discriminator-guided weights $(\mathcal{D}(y=1|v))$, and display them by showing the top-5 and bottom-5 instances, respectively. The weights are shown in the upper-left corner of each frame. We visualize eight video sequences that illustrate the following video quality degradation: \textbf{blurring}, \textbf{compression noise}, \textbf{lighting variation}, \textbf{video shot-cut}, \textbf{occlusion}, \textbf{detection failure}, \textbf{pose variation} and \textbf{mis-alignment} from the first to the last row.\label{fig:ytf_quality}}
\vspace{-5mm}
\end{figure*}

When feature restoration loss is used for training, we consider two types of discriminators since it introduces an additional data domain, namely a synthetic image domain, besides the existing two domains of image and video. First, we merge synthetically degraded images into video domain and the discriminator is still a two-way classifier (model E). Next, synthetic images are considered as their own domain, which leads to a three-way discriminator among image, synthetic image, and video (model F). In comparison to model C which is trained without adversarial loss, both models E and F significantly improve the recognition performance (e.g, from $93.82\%$ to $95.22\%$ or $95.00\%$ for model E and F, respectively). When frame-level features are aggregated by discriminator-guided fusion, the three-way model F improves its performance to $95.38\%$, which is highly competitive to the performance of previous state-of-the-art face recognition engines such as FaceNet~\cite{schroff2015facenet} ($95.12\%$), CenterFace~\cite{wen2016discriminative} ($94.9\%$), or CNN with different feature aggregation methods~\cite{Yang_2017_CVPR} (e.g., $95.20\%$ with average pooling) as shown in Table~\ref{tab:sota}. Note that the evaluation protocol of prior works is the same as our baseline model, but their networks are either much deeper or trained on significantly larger number of training images.
\begin{table}[htbp]
\centering
\caption{Comparison on the YTF dataset with other unsupervised domain adaptation methods and state-of-the-art methods.}
\vspace{0.04in}
\footnotesize
\begin{tabular}{c|c||c|c}
\hline
\multicolumn{2}{c||}{Unsupervised DA} & \multicolumn{2}{c}{SOTA (image-based)}\\
\hline
baseline & 93.78 & DeepFace~\cite{deepface2014} & 91.4 \\
PCA & 93.56 & FaceNet~\cite{schroff2015facenet} & 95.12 \\
CORAL~\cite{coral} & 94.50 & CenterFace~\cite{wen2016discriminative} & 94.9 \\
Ours (F) & \bf{95.38} & CNN+AvePool~\cite{Yang_2017_CVPR} & 95.20 \\\hline
\end{tabular}
\label{tab:sota}
\vspace{-0.2in}
\end{table}
\vspace{-0.3cm}
\paragraph{Discriminator-guided feature fusion.}
The proposed fusion strategy selectively adopts high-quality frames while discarding poorer ones in order to further improve the recognition accuracy. To reflect the quality of each frame, the feature fusion module aggregates all frames of a video using a weighted average of feature vectors based on the normalized likelihood as in~\eqref{eq:fuse}. We quantitatively and qualitatively evaluate the discriminator-guided fusion in Table~\ref{tab:ytf_summary} and Figure~\ref{fig:ytf_quality}. By applying $\mathcal{D}$ from the three-way network, the model F and the baseline model present consistent improvements. In contrast, the fusion strategy for the two-way network in model E (second sub-row) has marginal effect. This indicates that the three-way network learns a multi-class discriminator that can better distinguish among input sources.
\vspace{-0.3cm}
\paragraph{Qualitative visualization of guided fusion.}
In Figure \ref{fig:ytf_quality}, we demonstrate qualitative results for the feature fusion using the three-way discriminator scores. Each row shows the top and bottom scored frames. It is evident that high-quality frames score higher than low-quality ones. More importantly, we observe that the notions of quality are diverse and encompass factors of variation such as pose, blur, lighting and occlusions. This supports our hypothesis that there are several causes of domain gap between images and videos, so an adversarially trained discriminator is better than one that relies on enumerating all possible factors. Our analysis is reminiscent of that in \cite{Yang_2017_CVPR}, but we learn the quality of video frames in an unsupervised manner without identity labels, whereas \cite{Yang_2017_CVPR} utilizes identity labels to assign a higher score to a frame that contributes more to classification.
\vspace{-0.4cm}
\paragraph{Comparison with other unsupervised DA methods.}
We study the effectiveness of our proposed method in comparison to other works on unsupervised domain adaptation such as PCA feature transform or Correlation Alignment (CORAL)~\cite{coral} methods. We extract features from both still images and video frames using RFNet and apply PCA feature transform while retaining $90\%$ of the total variation. For CORAL, we calculate the mean ($\mu_{\mathcal{I}}, \mu_{\mathcal{V}}$) and covariance ($C_{\mathcal{I}}, C_{\mathcal{V}}$) of the features from two domains on the training set and transform features, for individual frames, as follows:
\begin{gather}
{\phi}(v) \leftarrow (\phi(v)-\mu_{\mathcal{V}}) C_{\mathcal{V}}^{-\frac{1}{2}}C_{\mathcal{I}}^{\frac{1}{2}} + \mu_{\mathcal{I}}
\end{gather}
The results in Table~\ref{tab:sota} show that simple feature transform method like PCA does not work well since it does not distinguish between two domains when computing the transformation matrix. On the other hand, CORAL demonstrates moderate improvement upon baseline by matching the first and second-order statistics of representations between two domains. Our method is also based on feature distribution matching between two domains through a discriminator, but allows learning a more complete transformation through end-to-end training of deep networks with a combination of losses, which results in more substantial improvements.

\vspace{-0.03in}
\subsection{IJB-A Dataset}\label{sec:ijb}
\vspace{-0.05in}
IJB-A~\cite{IJB} is a benchmark dataset for face recognition in the wild. It contains a mix of $5397$ still images and $20412$ video frames sampled from $2042$ videos of $500$ different subjects. The existence of video frames allows set-to-set comparison for verification, which opens up a new challenge for the face recognition problem. It is more challenging than LFW or YTF due to many factors of variation such as pose or facial expression as well as various image qualities.

There are 10 splits with about $30$k labeled images in each. Due to the small-sized training set, prior works pretrain the network using large-scale labeled datasets and use training set of each split for supervised fine-tuning. While we also use a pretrained face recognition engine for video face recognition, our network is fine-tuned on external video data without identity information and IJB-A is used for evaluation only. This is because the emphasis of our method is to use large-scale unlabeled video data to improve video face recognition rather than fine-tuning with manually labeled data. We first remove $379$ videos from the YTF training set as their identities overlap with those of IJB-A test set. We then utilize the YTF dataset without label information for network training and the unlabeled video fine-tuned network is evaluated on 10 splits of IJB-A test set.

We perform experiments in two settings. The 1:1 verification task compares genuine and impostor samples for one-to-one verification, while the 1:N task is to search samples against an enrolled gallery. We report the true acceptance rate (TAR) at different false acceptance rates (FAR) of $0.1$, $0.01$, and $0.001$ for verification and the rank-1, 5, 10 accuracy for identification in Table~\ref{tab:ijba_summary}. Compared to the baseline model, we observe slightly worse performance at FAR=$0.001$, but significant improvement on both FAR=$0.01$ and $0.1$ (e.g., 77.3\% to 82.8\% at FAR=$0.01$). Especially, when discriminator-guided feature fusion is used, the improvement becomes more significant. We also evaluate by following the protocol in \cite{wang2015face} as we observe non-negligible amount of localization errors while preprocessing. By removing poor quality images based on localization errors, we obtain much higher verification and identification results. However, the gap between our proposed model and the baseline is reduced, which we believe is due to low-quality images being mostly filtered out, so the baseline method can still work fine or even better. Compared to previous works, the performance of our proposed method is competitive under similar training and testing protocols~\cite{wang2015face,chen2016unconstrained}. Further improvement is expected by training a stronger base network on larger labeled image datasets \cite{Yang_2017_CVPR} or with other types of synthetic data augmentations such as pose variation with 3D synthesization \cite{masi2016we}. 

%%%%%%%%%%%%% CONCLUSION
\vspace{-0.05in}
\section{Conclusions}
\vspace{-0.05in}

Face recognition in videos presents unique challenges due to the paucity of large-scale datasets and several factors of variation that degrade frame quality. In this work, we address those challenges by proposing a novel feature-level domain adaptation approach that uses large-scale labeled still images and unlabeled video data. By distilling discriminative knowledge from a pretrained face recognition engine on labeled still images while adapting to video domain through synthetic data augmentation and domain adversarial training, we learn domain-invariant discriminative representations for video face recognition. Furthermore, we propose a discriminator-guided feature fusion method to effectively aggregate features from multiple frames and effectively rank them in accordance to their suitability for face recognition. We demonstrate the effectiveness of the proposed method for video face verification on the YTF and IJB-A benchmarks. Our future work will further exploit unsupervised domain adaptation to achieve continuous improvements through a growing collection of unlabeled videos.

\vspace{-0.05in}
\subsubsection*{Acknowledgments}
\vspace{-0.05in}
This work is supported in part by the NSF CAREER Grant {\#}1149783 and a gift from NEC Labs America. 

\newpage
{\small
	\bibliographystyle{ieee}
	\bibliography{videoface_arxiv}

\begin{thebibliography}{10}\itemsep=-1pt

\bibitem{chen2016unconstrained}
J.-C. Chen, V.~M. Patel, and R.~Chellappa.
\newblock Unconstrained face verification using deep cnn features.
\newblock IEEE, 2016.

\bibitem{chopra2005learning}
S.~Chopra, R.~Hadsell, and Y.~LeCun.
\newblock Learning a similarity metric discriminatively, with application to
  face verification.
\newblock In {\em CVPR}, 2005.

\bibitem{collobert:2011c}
R.~Collobert, K.~Kavukcuoglu, and C.~Farabet.
\newblock {Torch7}: A {Matlab}-like environment for machine learning.
\newblock In {\em BigLearn, NIPS Workshop}, 2011.

\bibitem{vface-cui2013}
Z.~Cui, W.~Li, D.~Xu, S.~Shan, and X.~Chen.
\newblock Fusing robust face region descriptors via multiple metric learning
  for face recognition in the wild.
\newblock In {\em CVPR}, 2013.

\bibitem{Cutting_1986}
J.~E. Cutting.
\newblock {\em Perception with an Eye for Motion}.
\newblock MIT Press, 1986.

\bibitem{vface-trunk2016}
C.~Ding and D.~Tao.
\newblock Trunk-branch ensemble convolutional neural networks for video-based
  face recognition.
\newblock {\em arXiv:1607.05427}, 2016.

\bibitem{uda-fernando2013}
B.~Fernando, A.~Habrard, M.~Sebban, and T.~Tuytelaars.
\newblock Unsupervised visual domain adaptation using subspace alignment.
\newblock In {\em ICCV}, 2013.

\bibitem{uda-fernando2015}
B.~Fernando, T.~Tommasi, and T.~Tuytelaars.
\newblock Joint cross-domain classification and subspace learning for
  unsupervised adaptation.
\newblock {\em Pattern Recoginition}, 2015.

\bibitem{ganin2015unsupervised}
Y.~Ganin and V.~Lempitsky.
\newblock Unsupervised domain adaptation by backpropagation.
\newblock In {\em ICML}, 2015.

\bibitem{Gibson_1979}
J.~Gibson.
\newblock {\em The Ecological Approach to Visual Perception}.
\newblock Houghton Mifflin, 1979.

\bibitem{goodfellow2014generative}
I.~Goodfellow, J.~Pouget-Abadie, M.~Mirza, B.~Xu, D.~Warde-Farley, S.~Ozair,
  A.~Courville, and Y.~Bengio.
\newblock Generative adversarial nets.
\newblock In {\em NIPS}, pages 2672--2680, 2014.

\bibitem{goodfellow2013maxout}
I.~Goodfellow, D.~Warde-farley, M.~Mirza, A.~Courville, and Y.~Bengio.
\newblock Maxout networks.
\newblock In {\em ICML}, 2013.

\bibitem{gupta2016cross}
S.~Gupta, J.~Hoffman, and J.~Malik.
\newblock Cross modal distillation for supervision transfer.
\newblock In {\em CVPR}, 2016.

\bibitem{he2016deep}
K.~He, X.~Zhang, S.~Ren, and J.~Sun.
\newblock Deep residual learning for image recognition.
\newblock In {\em CVPR}, 2016.

\bibitem{uda-distill}
G.~{Hinton}, O.~{Vinyals}, and J.~{Dean}.
\newblock {Distilling the Knowledge in a Neural Network}.
\newblock {\em CoRR}, abs/1503.02531, 2015.

\bibitem{LFWTech}
G.~B. Huang, M.~Ramesh, T.~Berg, and E.~Learned-Miller.
\newblock Labeled faces in the wild: A database for studying face recognition
  in unconstrained environments.
\newblock Technical report, University of Massachusetts, Amherst, 2007.

\bibitem{adam}
D.~P. Kingma and J.~Ba.
\newblock Adam: {A} method for stochastic optimization.
\newblock {\em CoRR}, abs/1412.6980, 2014.

\bibitem{IJB}
B.~F. Klare, B.~Klein, E.~Taborsky, A.~Blanton, J.~Cheney, K.~Allen,
  P.~Grother, A.~Mah, and A.~K. Jain.
\newblock Pushing the frontiers of unconstrained face detection and
  recognition: Iarpa janus benchmark a.
\newblock In {\em CVPR}, 2015.

\bibitem{Ledig_2017_CVPR}
C.~Ledig, L.~Theis, F.~Huszar, J.~Caballero, A.~Cunningham, A.~Acosta,
  A.~Aitken, A.~Tejani, J.~Totz, Z.~Wang, and W.~Shi.
\newblock Photo-realistic single image super-resolution using a generative
  adversarial network.
\newblock In {\em CVPR}, 2017.

\bibitem{vface-li2013}
H.~Li, G.~Hua, Z.~Lin, J.~Brandt, and J.~Yang.
\newblock Probabilistic elastic matching for pose variant face verification.
\newblock In {\em CVPR}, 2013.

\bibitem{vface-li2014}
H.~Li, G.~Hua, X.~Shen, Z.~Lin, and J.~Brandt.
\newblock Eigen-pep for video face recognition.
\newblock In {\em ACCV}, 2014.

\bibitem{liu2016coupled}
M.-Y. Liu and O.~Tuzel.
\newblock Coupled generative adversarial networks.
\newblock In {\em NIPS}, 2016.

\bibitem{uda-long2013}
M.~Long, J.~Wang, G.~Ding, J.~Sun, and P.~S. Yu.
\newblock Transfer feature learning with joint distribution adaptation.
\newblock In {\em ICCV}, 2013.

\bibitem{masi2016we}
I.~Masi, A.~T. Trần, T.~Hassner, J.~T. Leksut, and G.~Medioni.
\newblock Do we really need to collect millions of faces for effective face
  recognition?
\newblock In {\em ECCV}, 2016.

\bibitem{vggface15}
O.~M. Parkhi, A.~Vedaldi, and A.~Zisserman.
\newblock Deep face recognition.
\newblock 2015.

\bibitem{uda-saenko2010}
K.~Saenko, B.~Kulis, M.~Fritz, and T.~Darrell.
\newblock Adapting visual category models to new domains.
\newblock In {\em ECCV}, 2010.

\bibitem{schroff2015facenet}
F.~Schroff, D.~Kalenichenko, and J.~Philbin.
\newblock {FaceNet}: A unified embedding for face recognition and clustering.
\newblock In {\em CVPR}, 2015.

\bibitem{NIPS2016_6200}
K.~Sohn.
\newblock Improved deep metric learning with multi-class {N}-pair loss
  objective.
\newblock In {\em NIPS}. 2016.

\bibitem{coral}
B.~Sun, J.~Feng, and K.~Saenko.
\newblock Return of frustratingly easy domain adaptation.
\newblock In {\em AAAI}, 2016.

\bibitem{sun2014deep}
Y.~Sun, Y.~Chen, X.~Wang, and X.~Tang.
\newblock Deep learning face representation by joint
  identification-verification.
\newblock In {\em NIPS}, 2014.

\bibitem{gan-TaigmanPW16}
Y.~Taigman, A.~Polyak, and L.~Wolf.
\newblock Unsupervised cross-domain image generation.
\newblock {\em CoRR}, abs/1611.02200, 2016.

\bibitem{taigman2016unsupervised}
Y.~Taigman, A.~Polyak, and L.~Wolf.
\newblock Unsupervised cross-domain image generation.
\newblock 2017.

\bibitem{deepface2014}
Y.~Taigman, M.~Yang, M.~Ranzato, and L.~Wolf.
\newblock Deepface: Closing the gap to human-level performance in face
  verification.
\newblock In {\em CVPR}, 2014.

\bibitem{uda-TzengHZSD14}
E.~Tzeng, J.~Hoffman, N.~Zhang, K.~Saenko, and T.~Darrell.
\newblock Deep domain confusion: Maximizing for domain invariance.
\newblock {\em CoRR}, abs/1412.3474, 2014.

\bibitem{wang2015face}
D.~Wang, C.~Otto, and A.~K. Jain.
\newblock Face search at scale: 80 million gallery.
\newblock {\em arXiv preprint arXiv:1507.07242}, 2015.

\bibitem{Wang_Transformation}
X.~Wang, A.~Farhadi, and A.~Gupta.
\newblock Actions {\textasciitilde} transformations.
\newblock In {\em CVPR}, 2016.

\bibitem{wen2016discriminative}
Y.~Wen, K.~Zhang, Z.~Li, and Y.~Qiao.
\newblock A discriminative feature learning approach for deep face recognition.
\newblock In {\em ECCV}, pages 499--515. Springer, 2016.

\bibitem{vface-wolf2011}
L.~Wolf, T.~Hassner, and I.~Maoz.
\newblock Face recognition in unconstrained videos with matched background
  similarity.
\newblock In {\em CVPR}, 2011.

\bibitem{vface-Wolf2013}
L.~Wolf and N.~Levy.
\newblock The svm-minus similarity score for video face recognition.
\newblock In {\em CVPR}, 2013.

\bibitem{Yang_2017_CVPR}
J.~Yang, P.~Ren, D.~Zhang, D.~Chen, F.~Wen, H.~Li, and G.~Hua.
\newblock Neural aggregation network for video face recognition.
\newblock In {\em CVPR}, 2017.

\bibitem{YD2014}
D.~Yi, Z.~Lei, S.~Liao, and S.~Z. Li.
\newblock Learning face representation from scratch.
\newblock {\em CoRR}, abs/1411.7923, 2014.

\bibitem{yoo2016pixel}
D.~Yoo, N.~Kim, S.~Park, A.~S. Paek, and I.~S. Kweon.
\newblock Pixel-level domain transfer.
\newblock In {\em ECCV}, 2016.

\bibitem{yu2016deep}
X.~Yu, F.~Zhou, and M.~Chandraker.
\newblock Deep deformation network for object landmark localization.
\newblock In {\em ECCV}, 2016.

\bibitem{zhou2016view}
T.~Zhou, S.~Tulsiani, W.~Sun, J.~Malik, and A.~A. Efros.
\newblock View synthesis by appearance flow.
\newblock In {\em ECCV}, 2016.

\end{thebibliography}
}

\appendix
\renewcommand{\thesection}{S\arabic{section}}   
\renewcommand{\thetable}{S\arabic{table}}   
\renewcommand{\thefigure}{S\arabic{figure}}
\renewcommand{\theequation}{S\arabic{equation}}

%%%%%%%%%% SUPP: NETARCH
\section{Implementation Details}
\label{sec:netarch}

In this section, we present detailed network architecture as well as implementation details, for reproducible research. The network architecture for the reference network (RFNet) and video domain-adapted network (VDNet) are equivalent, as described in Table~\ref{tab:casia_maxout}. The network architecture is mainly motivated from~\cite{YD2014}, but we replace ReLU with maxout units at every other convolution layer to further improve the performance while maintaining the same number of neurons at each layer.

The RFNet is pretrained on the CASIA webface database~\cite{YD2014}, which includes 494414 images from 10575 identities from the Internet, using the same training setup~\cite{NIPS2016_6200}. The implementation is based on Torch~\cite{collobert:2011c} with $N=1080$ (that is, number of examples per batch is set to $2160$) for N-pair loss. The N-pair loss, which pushes (N-1) negative examples at the same time while pulling a single positive example, is used on 8 GPUs for training. The VDNet is initialized with the RFNet followed by a discriminator composed of two fully connected layers (160, 3) followed by a ReLU on top of $320$-dimensional output feature of VDNet. The VDNet is trained with the following objective function:
\begin{equation}
\mathcal{L} = \mathcal{L}_{\text{FM}} + \alpha\mathcal{L}_{\text{FR}} + \beta\mathcal{L}_{\text{IC}} + \gamma\mathcal{L}_{\text{Adv}}\label{eq:objective-all-supp}
\end{equation}
where the forms of the loss functions are described in the main paper and we set $\alpha=\beta=\gamma=1$ for all our experiments. The learning rate is set to $0.0003$ with a default setting of the Adam optimizer~\cite{adam} (for example, $\beta_1=0.9$, $\beta_2=0.999$). The network is trained for $1500$ iterations with batch size of $512$, where we allocate $256$ examples from the image domain and remaining $256$ examples from the video domain for each mini batch.

%% model definition
\begin{table}[htbp]
\centering
\caption{Network architecture for RFNet and VDNet. The network is composed of 10 layers of 3$\times$3 convolution layers followed by either ReLU or maxout units~\cite{goodfellow2013maxout}. The volumetric max pooling (Vmax pooling) extends (spatial) max pooling to input channels and can be used to model maxout units. \label{tab:casia_maxout}}
\vspace{0.05in}
\small{
\begin{tabular}{c|c|c}
\hline
operation & kernel & output size\\
\hline
Conv1-1 + ReLU & 3$\times$3 & 100$\times$100$\times$32\\
\hline
Conv1-2 & 3$\times$3 & 100$\times$100$\times$128\\
\hline
Vmax pooling & {2$\times$2$\times$2} & {50$\times$50$\times$64}\\
\hline
Conv2-1 + ReLU & 3$\times$3 & 50$\times$50$\times$64\\
\hline
Conv2-2 & 3$\times$3 & 50$\times$50$\times$256\\
\hline
Vmax pooling & {2$\times$2$\times$2} & {25$\times$25$\times$128}\\
\hline
Conv3-1 + ReLU & 3$\times$3 & 25$\times$25$\times$96\\
\hline
Conv3-2 & 3$\times$3 & 25$\times$25$\times$384\\
\hline
Vmax pooling & {2$\times$2$\times$2} & {13$\times$13$\times$192}\\
\hline
Conv4-1 + ReLU & 3$\times$3 & 13$\times$13$\times$128\\
\hline
Conv4-2 & 3$\times$3 & 13$\times$13$\times$512\\
\hline
Vmax pooling & {2$\times$2$\times$2} & {7$\times$7$\times$256}\\
\hline
Conv5-1 + ReLU & 3$\times$3 & 7$\times$7$\times$160\\
\hline
Conv5-2 & 3$\times$3 & 7$\times$7$\times$320\\
\hline
Average pooling & 7$\times$7 & 1$\times$1$\times$320\\
\hline
\end{tabular}
}
\end{table}

\begin{figure*}[t!]
	\centering
	\includegraphics[width=0.98\textwidth]{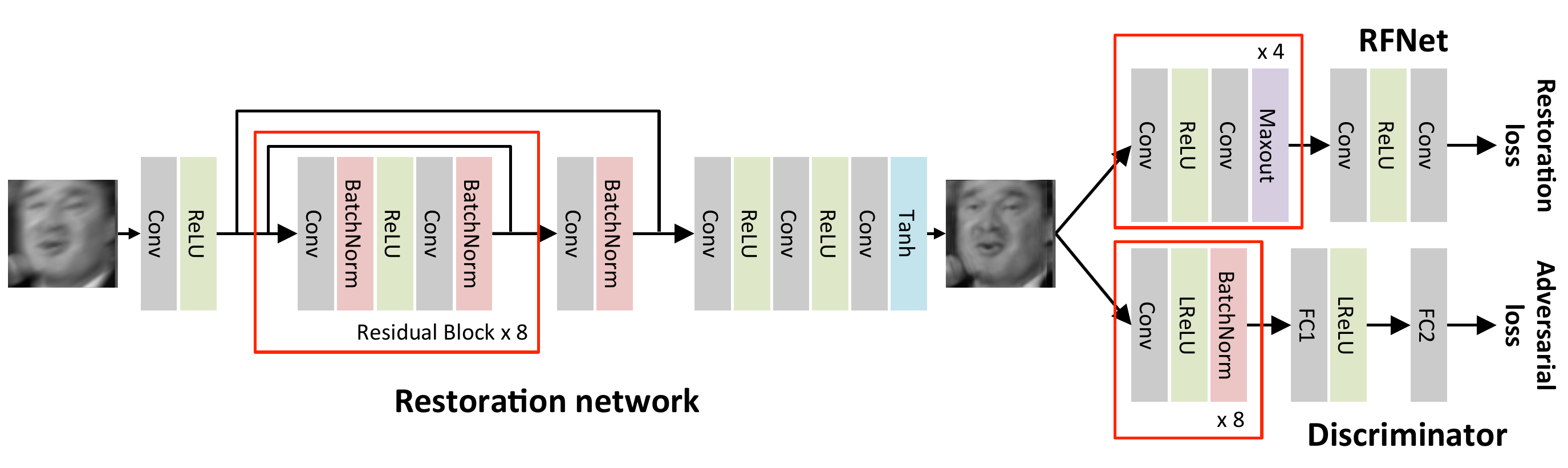}
	\caption{Illustration of pixel-space restoration and adaptation framework. The restoration network is composed of 8 residual blocks and a shortcut connection as described in Table~\ref{tab:restrnet}. The network architectures of RFNet and discriminator are provided in Table~\ref{tab:casia_maxout} and~\ref{tab:disc}, respectively. The input for restoration network is either synthetically degraded images (for IRResNet or IRGAN) or video frames (for VRGAN) and the output is fed into either RFNet for feature restoration loss with restored images or discriminator for adversarial loss with both restored images and video frames.\label{fig:restrnet_diagram}}
\end{figure*}

\begin{table}[htbp]
\centering
\caption{Architecture for image or video pixel-space restoration network. The network is composed of 8 residual blocks and few more convolution layers before and after a series of residual blocks. There also exists a shortcut connection where the output of C1 and C2 are added before fed into C3. \label{tab:restrnet}}
\vspace{0.05in}
\small{
\begin{tabular}{c|c|c|c}
\hline
name & operation & kernel & output size\\
\hline
C1 & Conv + ReLU & 3$\times$3 & 100$\times$100$\times$32\\
\hline
\multirow{2}{*}{Res1--8} & Conv + BN + ReLU & \multirow{2}{*}{3$\times$3} & \multirow{2}{*}{100$\times$100$\times$32}\\
 & + Conv + BN &  & \\
\hline
C2 & Conv + BN & 3$\times$3 & 100$\times$100$\times$32\\
\hline
C3 & Conv + ReLU & 3$\times$3 & 100$\times$100$\times$32\\
\hline
C4 & Conv + ReLU & 3$\times$3 & 100$\times$100$\times$32\\
\hline
C5 & Conv + Tanh & 3$\times$3 & 100$\times$100$\times$1\\
\hline
\end{tabular}
}
\end{table}

\begin{table}[htbp]
\centering
\caption{Network architecture for discriminator. The network is composed of 8 layers of 3$\times$3 convolution layers followed by by Leaky ReLU (LReLU) and Batch Normalization (BN) layers and 2 fully-connected (FC) layers whose final output is either 2 for IRGAN or 3 for VRGAN. \label{tab:disc}}
\vspace{0.05in}
\small{
\begin{tabular}{c|c|c}
\hline
operation & kernel, stride & output size\\
\hline
Conv1-1 + LReLU & 3$\times$3, 1 & 100$\times$100$\times$64\\
\hline
Conv1-2 + LReLU + BN & 3$\times$3, 2 & 50$\times$50$\times$64\\
\hline
Conv2-1 + LReLU + BN & 3$\times$3, 1 & 50$\times$50$\times$128\\
\hline
Conv2-2 + LReLU + BN & 3$\times$3, 2 & 25$\times$25$\times$128\\
\hline
Conv3-1 + LReLU + BN & 3$\times$3, 1 & 25$\times$25$\times$256\\
\hline
Conv3-2 + LReLU + BN & 3$\times$3, 2 & 13$\times$13$\times$256\\
\hline
Conv4-1 + LReLU + BN & 3$\times$3, 1 & 13$\times$13$\times$512\\
\hline
Conv4-2 + LReLU + BN & 3$\times$3, 2 & 7$\times$7$\times$512\\
\hline
FC1 + LReLU & -- & 1024\\
\hline
\multirow{2}{*}{FC2} & \multirow{2}{*}{--} & 2 (IRGAN)\\
&  & 3 (VRGAN)\\
\hline
\end{tabular}
}
\end{table}

\section{Ablation Study}
\label{sec:control_exp}
In this section, we present further results for different design choices of the proposed algorithm. In particular, we consider the alternative of pixel-space image restoration and study the effect of scale of unlabeled video training data. 

\begin{table}[t]
\centering
\caption{Face verification accuracy on the degraded LFW dataset. The baseline network (RFNet) is evaluated on both degraded and original ($^{\dagger}$) test set. We run evaluation on degraded test images for $10$ times with different random seeds and report the mean accuracy. \label{tab:lfw_summary}}
\vspace{0.03in}
\footnotesize
\begin{tabular}{c|c|c|c|c|c}
\hline
Model & baseline & IRResNet & IRGAN & VRGAN & VDNet-F\\
\hline
VRF & 88.68 & \multirow{2}{*}{92.69} & \multirow{2}{*}{92.38} & \multirow{2}{*}{92.41} & \multirow{2}{*}{93.72}\\
\cline{1-2}
VRF$^{\dagger}$ & 98.85 & & & & \\
\hline
\end{tabular}
\end{table}

\begin{table*}[t]
\centering
\caption{Video face verification accuracy and standard error on the YTF database with different image- and video-restoration networks. The verification accuracy averaged over 10 folds and corresponding standard error are reported. The best performer and those with overlapping standard error are boldfaced. \label{tab:ytf_summary_vrgan}}
\vspace{0.04in}
\footnotesize
\begin{tabular}{c|c||c|c|c|c|c}
\hline
Model & fusion & 1 (fr/vid) & 5 (fr/vid) & 20 (fr/vid) & 50 (fr/vid) & all\\
\hline
baseline & -- & 91.12{\footnotesize$\pm 0.318$} & 93.17{\footnotesize$\pm 0.371$} & 93.62{\footnotesize$\pm 0.430$} & 93.74{\footnotesize$\pm 0.443$} & 93.78{\footnotesize$\pm 0.498$}\\
IRResNet & -- & 90.40{\footnotesize$\pm 0.366$} & 92.59{\footnotesize$\pm 0.388$} & 93.13{\footnotesize$\pm 0.405$} & 93.15{\footnotesize$\pm 0.416$} & 93.26{\footnotesize$\pm 0.400$}\\
IRGAN & -- & 90.67{\footnotesize$\pm 0.314$} & 92.88{\footnotesize$\pm 0.369$} & 93.25{\footnotesize$\pm 0.389$} & 93.24{\footnotesize$\pm 0.368$} & 93.22{\footnotesize$\pm 0.383$}\\
VRGAN & -- & 90.77{\footnotesize$\pm 0.346$} & 92.93{\footnotesize$\pm 0.402$} & 93.41{\footnotesize$\pm 0.429$} & 93.46{\footnotesize$\pm 0.410$} & 93.62{\footnotesize$\pm 0.439$}\\
\hline
\multirow{2}{*}{F (ours)} & -- & 92.17{\footnotesize$\pm 0.353$} & 94.44{\footnotesize$\pm 0.343$} & \bf{94.90}{\footnotesize$\pm 0.345$} & \bf{94.98}{\footnotesize$\pm 0.354$} & \bf{95.00}{\footnotesize$\pm 0.415$}\\
& \checkmark & -- & 94.52{\footnotesize$\pm 0.356$} & \bf{95.01}{\footnotesize$\pm 0.352$} & \bf{95.15}{\footnotesize$\pm 0.370$} & \bf{95.38}{\footnotesize$\pm 0.310$}\\
\hline
\end{tabular}
\end{table*}

\begin{table*}[t]
\centering
\caption{Video face verification accuracy and standard error on the YTF database with different number of unlabeled videos for domain-adversarial training. The verification accuracy averaged over 10 folds and corresponding standard error are reported. VDNet model F is used for experiments, where all four losses including feature matching, feature restoration, image classification, as well as domain adversarial losses, are used. The best performer and those with overlapping standard error are boldfaced. \label{tab:ytf_summary_scale}}
\vspace{0.04in}
\footnotesize
\begin{tabular}{c|c||c|c|c|c|c}
\hline
$\#$ videos & fusion & 1 (fr/vid) & 5 (fr/vid) & 20 (fr/vid) & 50 (fr/vid) & all\\
\hline
\multirow{2}{*}{10} & -- & 91.54{\footnotesize$\pm 0.339$} & 93.62{\footnotesize$\pm 0.338$} & 94.05{\footnotesize$\pm 0.350$} & 94.17{\footnotesize$\pm 0.390$} & 94.16{\footnotesize$\pm 0.369$}\\
& \checkmark & -- & 93.63{\footnotesize$\pm 0.365$} & 94.12{\footnotesize$\pm 0.377$} & 94.22{\footnotesize$\pm 0.386$} & 94.22{\footnotesize$\pm 0.381$}\\
\hline
\multirow{2}{*}{25} & -- & 91.80{\footnotesize$\pm 0.337$} & 93.84{\footnotesize$\pm 0.320$} & 94.24{\footnotesize$\pm 0.328$} & 94.38{\footnotesize$\pm 0.321$} & 94.46{\footnotesize$\pm 0.323$}\\
& \checkmark & -- & 93.90{\footnotesize$\pm 0.331$} & 94.29{\footnotesize$\pm 0.338$} & 94.40{\footnotesize$\pm 0.334$} & 94.56{\footnotesize$\pm 0.333$}\\
\hline
\multirow{2}{*}{100} & -- & 92.34{\footnotesize$\pm 0.289$} & 94.42{\footnotesize$\pm 0.348$} & \bf{94.78}{\footnotesize$\pm 0.379$} & \bf{94.82}{\footnotesize$\pm 0.385$} & \bf{94.78}{\footnotesize$\pm 0.363$}\\
& \checkmark & -- & 94.50{\footnotesize$\pm 0.331$} & \bf{94.85}{\footnotesize$\pm 0.388$} & \bf{94.87}{\footnotesize$\pm 0.403$} & \bf{95.00}{\footnotesize$\pm 0.417$}\\
\hline
\multirow{2}{*}{250} & -- & 92.03{\footnotesize$\pm 0.348$} & 94.21{\footnotesize$\pm 0.342$} & 94.70{\footnotesize$\pm 0.314$} & 94.73{\footnotesize$\pm 0.324$} & \bf{94.90}{\footnotesize$\pm 0.291$}\\
& \checkmark & -- & 94.23{\footnotesize$\pm 0.348$} & 94.70{\footnotesize$\pm 0.322$} & \bf{94.81}{\footnotesize$\pm 0.313$} & \bf{94.98}{\footnotesize$\pm 0.323$}\\
\hline
all & -- & 92.17{\footnotesize$\pm 0.353$} & 94.44{\footnotesize$\pm 0.343$} & \bf{94.90}{\footnotesize$\pm 0.345$} & \bf{94.98}{\footnotesize$\pm 0.354$} & \bf{95.00}{\footnotesize$\pm 0.415$}\\
($\sim$2780) & \checkmark & -- & 94.52{\footnotesize$\pm 0.356$} & \bf{95.01}{\footnotesize$\pm 0.352$} & \bf{95.15}{\footnotesize$\pm 0.370$} & \bf{95.38}{\footnotesize$\pm 0.310$}\\
\hline
\end{tabular}
\end{table*}

\subsection{Pixel-space Restoration and Adaptation}
\label{sec:img_restr}
The focus of our paper is feature-level domain adaptation. In Section 3.3 of the main paper, lines 368--371, we state that this is preferable over pixel-level alternatives. To illustrate this further, we now compare the performance of our proposed method to several baselines on pixel-space restoration and domain adaptation.

The pixel-space restoration applies a similar combination of loss strategies, but shifts the restored domain from feature to image pixels. Instead of training VDNet, we use RFNet for feature restoration loss and a discriminator for domain adversarial loss on top of a ``restored'' input image, with an additionally trained image restoration network. Based on the single-image super-resolution generative adversarial network (SRGAN)~\cite{Ledig_2017_CVPR}, we train several image restoration networks for face images, which we call image-restoration ResNet (IRResNet), image-restoration GAN (IRGAN) and video-restoration GAN (VRGAN), as follows:
\begin{tight_itemize}
\item {\bf IRResNet}: An image restoration network based on very deep residual network~\cite{he2016deep}, trained with feature restoration loss guided by pretrained face recognition engine (RFNet) on synthetically degraded images. 
\item {\bf IRGAN}: An image restoration network based on very deep residual network trained with feature restoration loss as well as discriminator loss on synthetically degraded images.
\item {\bf VRGAN}: A video restoration network based on very deep residual network trained with feature restoration loss as well as discriminator loss on both synthetically degraded images and video frames.
\end{tight_itemize}
The pixel-space restoration models are illustrated in Figure~\ref{fig:restrnet_diagram}. The network architectures for the pixel-space restoration network and discriminator are summarized in Tables \ref{tab:restrnet} and \ref{tab:disc}, respectively. We use grayscale images for restoration networks as our RFNet accepts grayscaled images as input.\footnote{Although we use grayscale images as input and output of the restoration network, one can construct the restoration network for RGB images with an additional differentiable layer that converts RGB images into grayscale images before feeding into RFNet.} We use Adam optimizer with a learning rate of $0.0003$. Compared to the training of VDNet, we reduce the batch size to 96, where we allocate 48 examples from the image domain and another 48 examples from the video domain for each mini batch. This is due to additional CNNs such as restoration network and discriminator. We increase the number of iterations to $30000$ due to slower convergence.

\begin{figure*}[t!]
\centering
\includegraphics[width=0.98\textwidth]{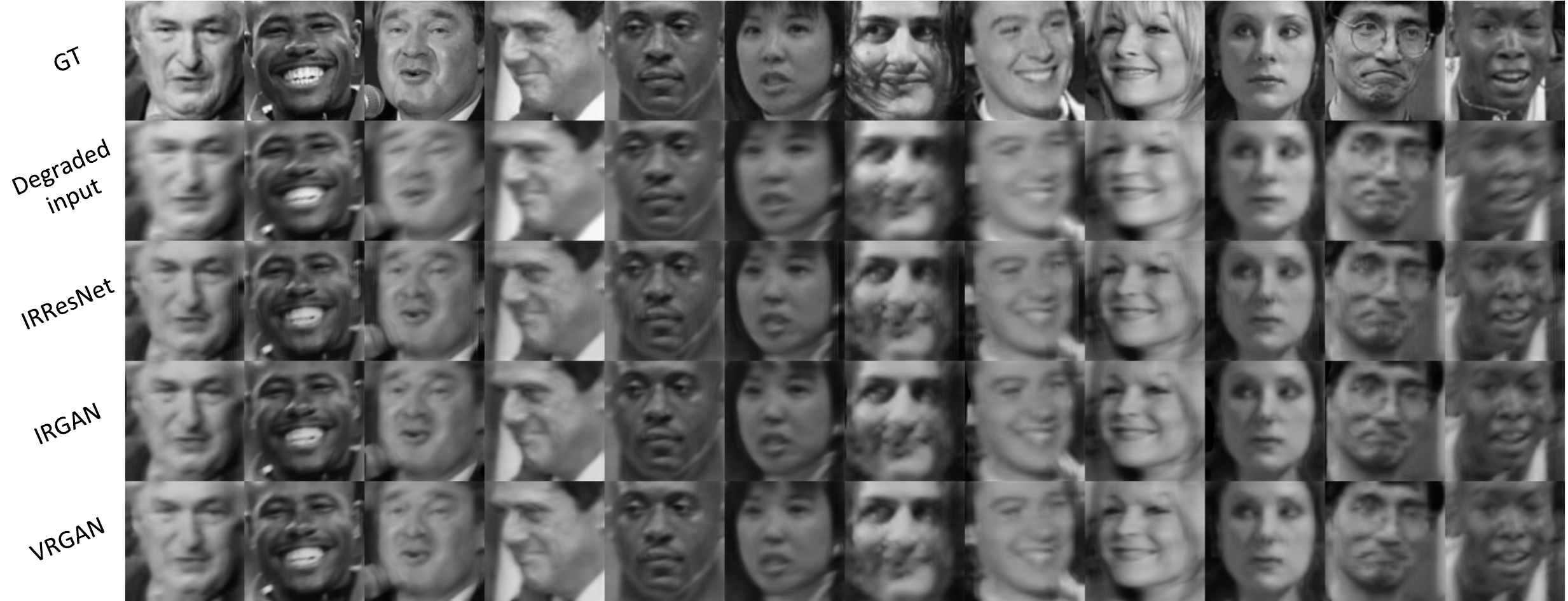}
\vspace{0.04in}
\caption{Image restoration on synthetically degraded LFW test set. From top to bottom, we visualize ground truth images, synthetically degraded images, and restored images with IRResNet, IRGAN, and VRGAN, respectively. \label{fig:vis_synth_restr}}
\end{figure*}

For fair comparisons, we apply the same set of random noise processes to generate synthetically degraded images for training, as described in Section~\ref{sec:restoration}. We note that other approaches might be used for image restoration, including ones that rely on further supervision through specification of a restoration target within a video sequence. But our baselines based on synthetically degraded images are reasonable in a setting comparable to VDNet that uses only unlabeled videos and produce visually good results. In particular, we visualize the image restoration results on synthetically degraded images of the LFW test set in Figure~\ref{fig:vis_synth_restr} and note that each baseline produces qualitatively reasonable outputs.

For quantitative validation of IRResNet, IRGAN and VRGAN, we evaluate the face verification performance on synthetically degraded Labeled Faces in the Wild (LFW)~\cite{LFWTech} dataset, where we apply the same set of image degradation kernels as used in the training with images of the LFW test set. The results are summarized in Table~\ref{tab:lfw_summary}. The image restoration methods effectively improve the performance from $88.68\%$ to $92.69\%$, $92.38\%$, and $92.41\%$ with IRResNet, IRGAN, and VRGAN, respectively. On this test set, adversarial training does not improve the verification performance since it aims to perform global distribution adaptation, thereby losing discriminative information that can be directly learned from the feature restoration loss defined between corresponding synthetic and original images. However, none of the image restoration models are as effective as the feature-level restoration model (model F), which achieves $93.72\%$.

We further evaluate the verification performance on the YouTube Face database (YTF) whose results are summarized in Table~\ref{tab:ytf_summary_vrgan}. Different from the results on the synthetically degraded LFW dataset, there is no performance gain when evaluated on YTF dataset with image and video restoration networks. In addition, we observe significant performance drops when trained only on the synthetic image database. Training VRGAN with an additional domain adversarial loss for video data improves the performance on video face verification over restoration models trained only with synthetically degraded images, but the improvement is not as significant as we have observed from the VDNet experiments.

In summary, it is evident that aligning distributions in pixel space is more difficult and there is a clear advantage of feature-level domain adaptation, especially when our ultimate goal is to improve the performance of high-level tasks such as classification.

\subsection{Number of Unlabeled Videos}
We perform controlled experiments with different number of unlabeled videos at training. Specifically, we utilize randomly selected $10$, $25$, $100$ and $250$ videos of YTF database for training. As shown in Table~\ref{tab:ytf_summary_scale}, we observe a general trend that the more video data is used for training, the higher verification accuracy we obtain. For example, if we use only $10$ videos for training, we obtain $94.22\%$ accuracy which is far lower than the best accuracy of $95.38\%$ achieved using all videos for training, which is approximately $2780$ unlabeled videos for each training fold. It is also worth noting that even using a very small number of unlabeled videos for training already improves the performance over other models that do not use domain adversarial training, such as model B ($93.94\%$) or model C ($93.82\%$). But at the same time, the margin of improvement gets smaller as we include more unlabeled videos for training.

Our paper demonstrates initial success with utilizing unlabeled video data for video face recognition, but interesting further problems remain. A promising direction of future is to explore ways to better utilize the fact that unlabeled videos are easier to acquire than labeled ones, translating to steady improvements in recognition performance as progressively larger scales of unlabeled data become available.

\section{Qualitative Visualization of Guided Fusion}
\begin{figure*}[t!]
\centering
\begin{tabular}{@{\hspace{0mm}}c@{\hspace{0mm}}c}
\centering
\includegraphics[width=0.48\textwidth]{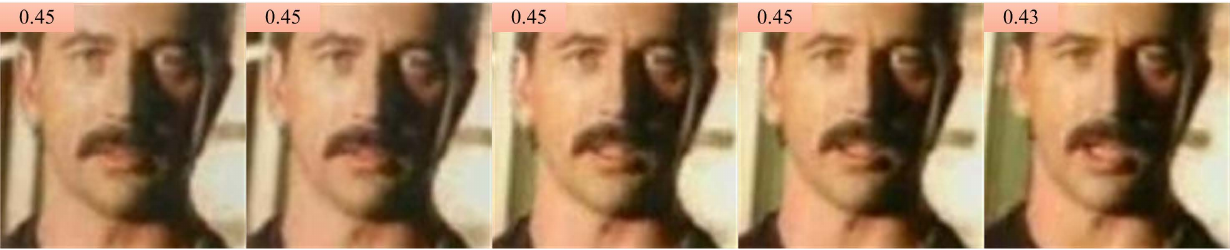}&\quad
\includegraphics[width=0.48\textwidth]{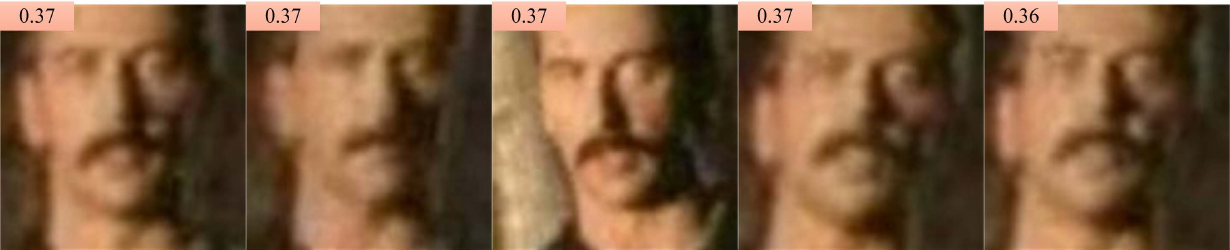}\\
\includegraphics[width=0.48\textwidth]{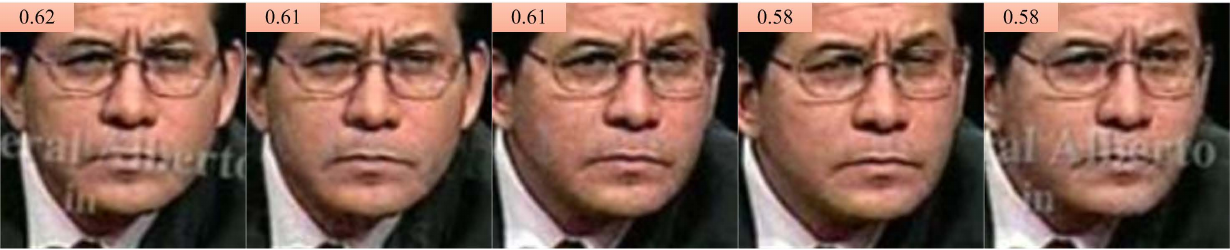}&\quad
\includegraphics[width=0.48\textwidth]{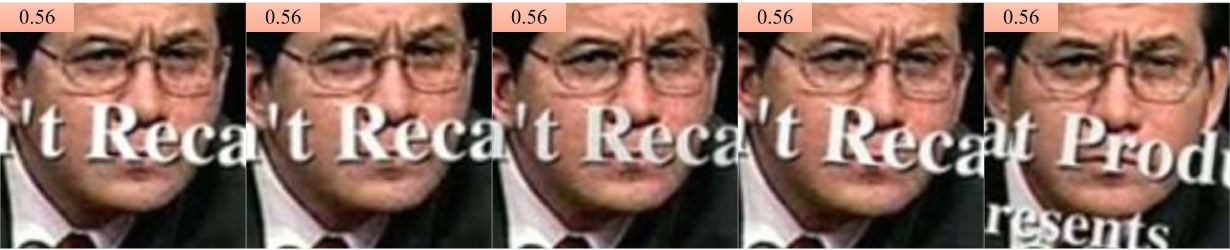}\\
\includegraphics[width=0.48\textwidth]{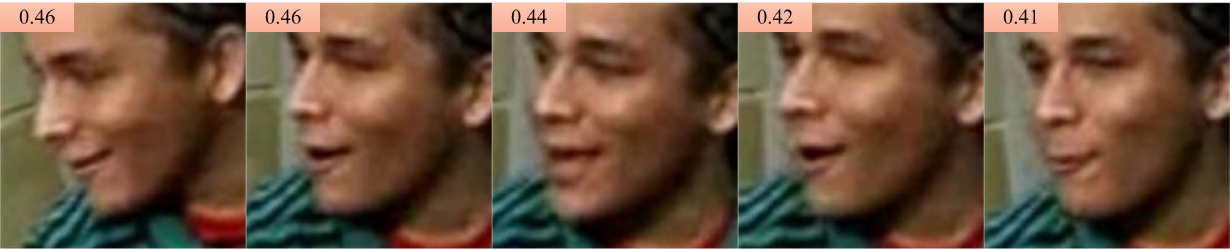}&\quad
\includegraphics[width=0.48\textwidth]{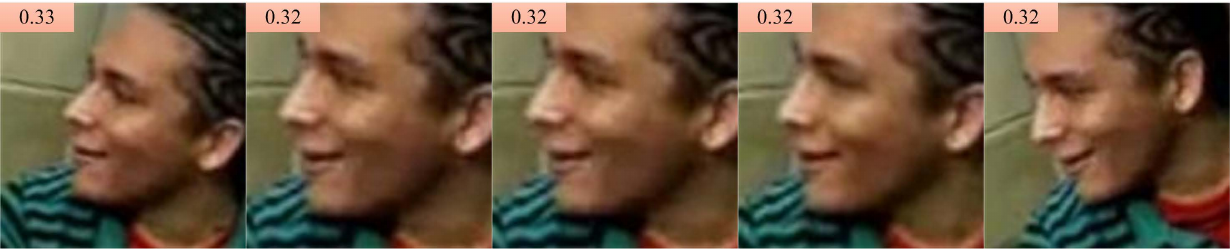}\\
\includegraphics[width=0.48\textwidth]{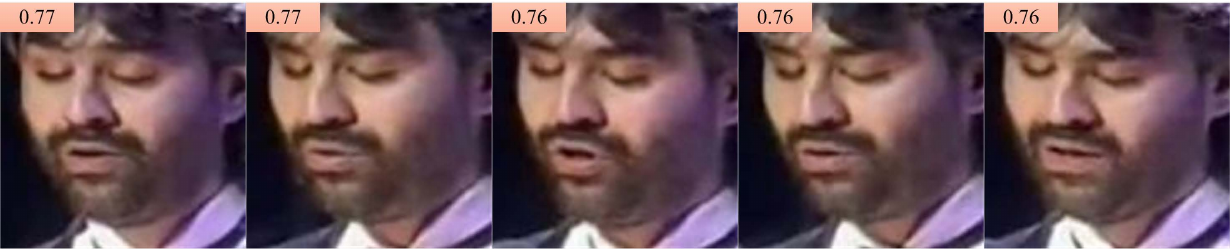}&\quad
\includegraphics[width=0.48\textwidth]{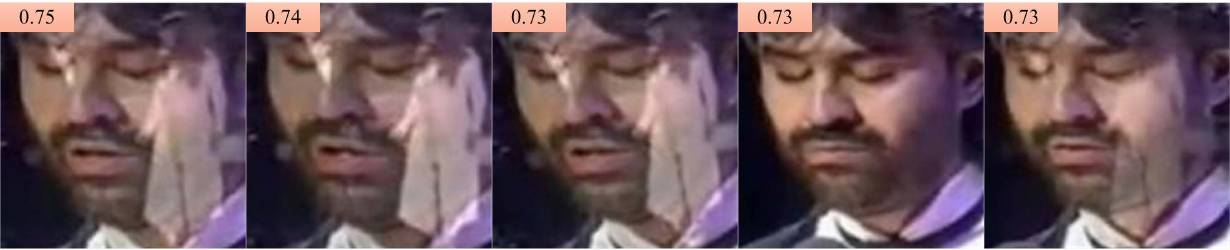}\\
\includegraphics[width=0.48\textwidth]{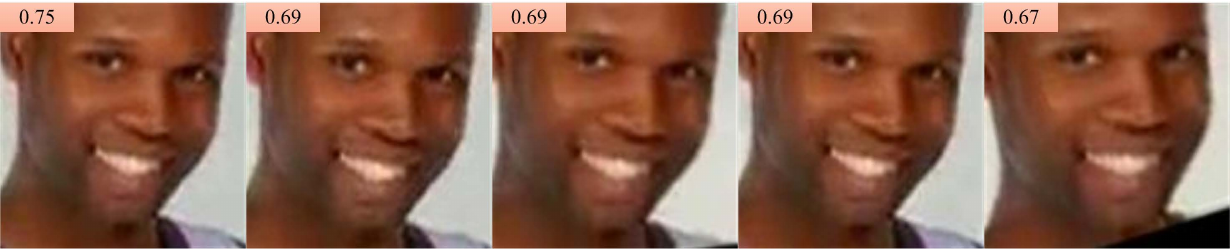}&\quad
\includegraphics[width=0.48\textwidth]{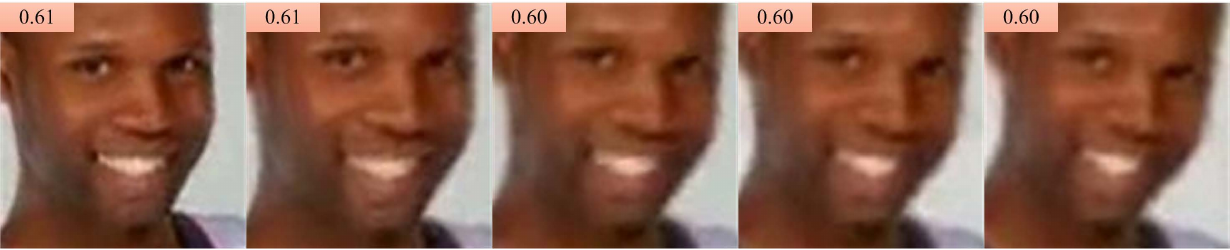}\\
\includegraphics[width=0.48\textwidth]{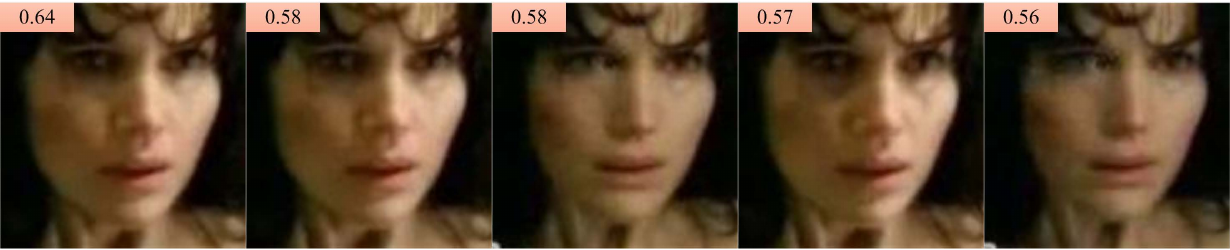}&\quad
\includegraphics[width=0.48\textwidth]{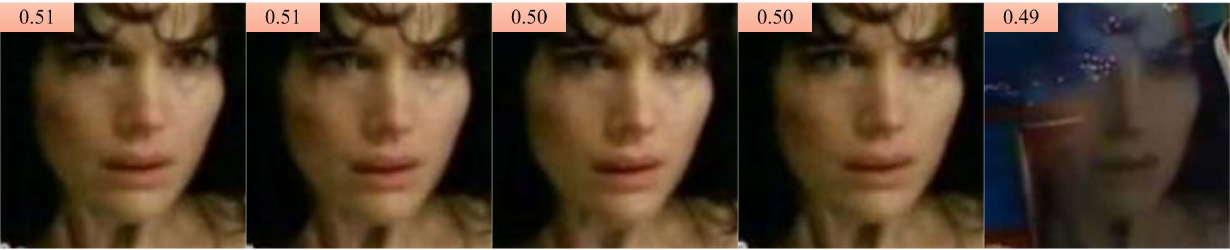}\\
\includegraphics[width=0.48\textwidth]{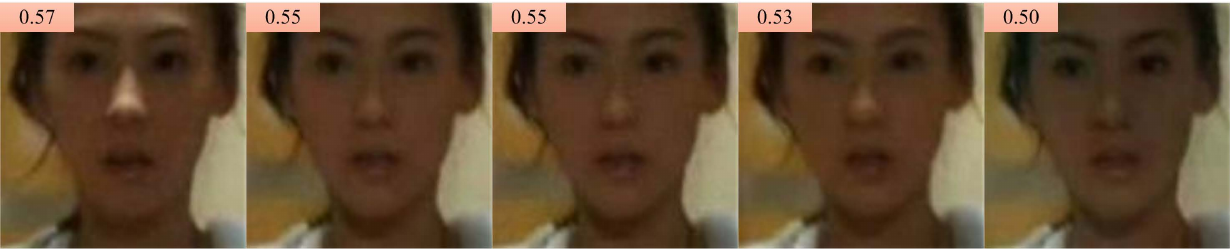}&\quad
\includegraphics[width=0.48\textwidth]{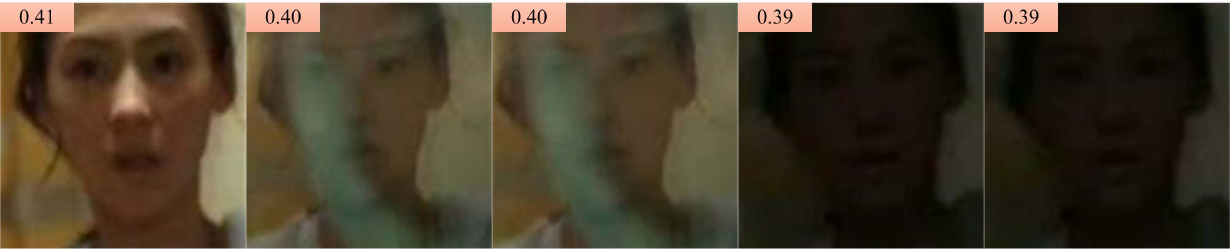}\\
\includegraphics[width=0.48\textwidth]{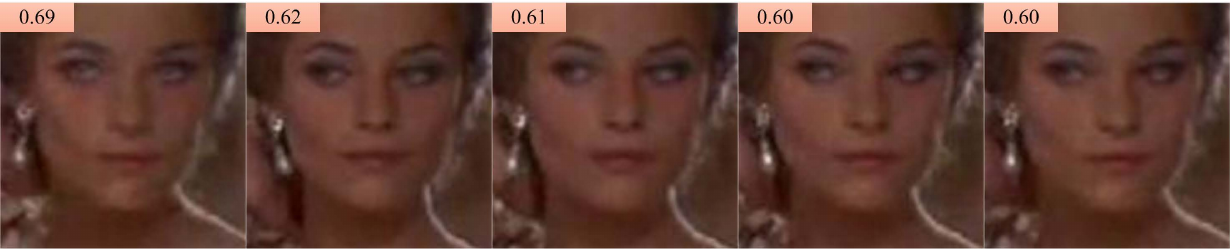}&\quad
\includegraphics[width=0.48\textwidth]{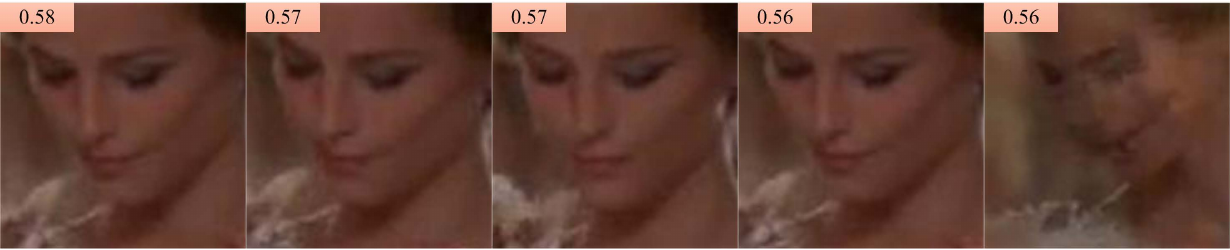}\\
\includegraphics[width=0.48\textwidth]{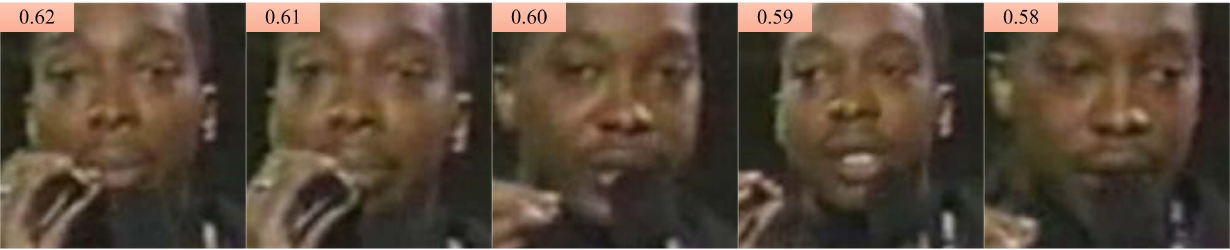}&\quad
\includegraphics[width=0.48\textwidth]{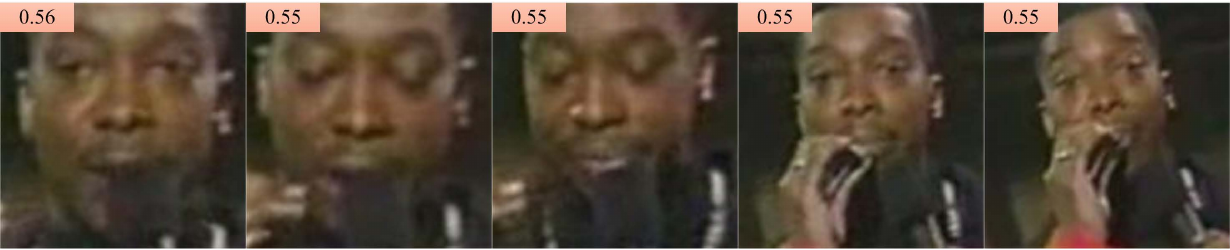}\\
\includegraphics[width=0.48\textwidth]{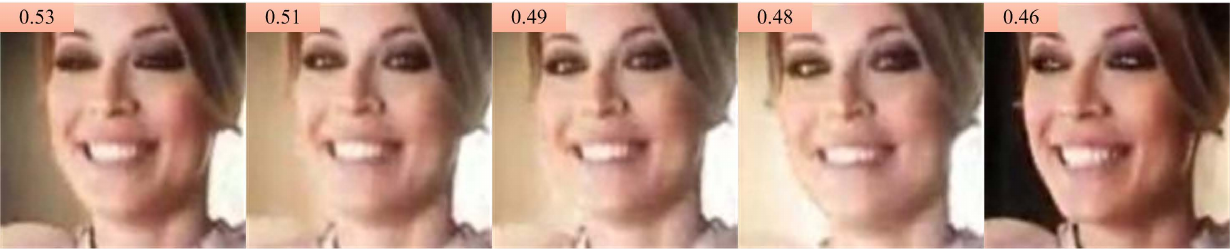}&\quad
\includegraphics[width=0.48\textwidth]{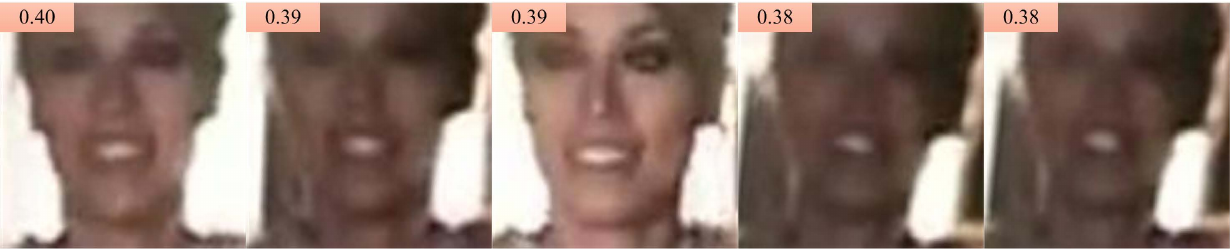}\\
{ (a) top-5}& {(b) bottom-5 } \\
\end{tabular}
\vspace{0.04in}
\caption{We sort the frames within a sequence in a descending order with respect to the confidence score of three-way discriminator $(\mathcal{D}(y=1|v))$, and display them by showing the top-5 and bottom-5 instances, respectively. The weights are shown in the upper-left corner of each frame.\label{fig:ytf_quality_ext}}
\vspace{-5mm}
\end{figure*}

We present further examples for the qualitative visualization of three-way domain discriminator. Similar to Figure~\ref{fig:ytf_quality}, we sort and show ten additional video clips with the confidence score of discriminator in Figure~\ref{fig:ytf_quality_ext}. As in the main paper, we observe that the discriminator ranks the video frames in a reasonable order, encompassing variations along several meaningful factors.

\end{document}